\documentclass{article}

\usepackage[final]{corl_2026} 
\usepackage{colortbl}
\usepackage{float}
\usepackage{xcolor}
\usepackage{placeins}
\usepackage{multirow}
\usepackage{algorithm}
\usepackage{amsmath}
\usepackage{amssymb}
\usepackage{bbm}
\usepackage{algpseudocode}
\usepackage{graphicx}
\usepackage{booktabs} 
\usepackage[percent]{overpic}
\usepackage{wrapfig}
\usepackage{caption}
\usepackage{subcaption}
\usepackage[percent]{overpic}
\usepackage{tikz}

\title{NTR: Neural Token Reconstruction for Scene Token Bottleneck in End-to-End Driving}

\author{
  \textbf{Jiahui Li}$^{1,2,*,\S}$ \quad
  \textbf{Jiawei Sun}$^{1,2,*,\dagger}$ \quad
  \textbf{Zixiang Ren}$^{1}$ \quad
  \textbf{Ming Liu}$^{1}$ \quad
  \textbf{Jiamin Shi}$^{1}$ \\
  \textbf{Ruiteng Zhao}$^{1}$ \quad
  \textbf{Zhiyang Liu}$^{1}$ \quad
  \textbf{Liying Liu}$^{2}$ \quad
  \textbf{Zuoguan Wang}$^{2}$ \quad
  \textbf{Kaidi Yang}$^{1,\ddagger}$ \\[0.4em]
  {\normalfont
    $^{1}$National University of Singapore, Singapore \quad
    $^{2}$Black Sesame Technologies
  } \\[0.2em]
  {\normalfont\small
    \texttt{\{jiahui.li, sunjiawei, e1554260, e1583192\}@u.nus.edu}
  } \\
  {\normalfont\small
    \texttt{\{e1704770, ruiteng, zhiyang\}@u.nus.edu} \quad
    \texttt{ykaidi@nus.edu.sg}
  } \\
  {\normalfont\small
    \texttt{\{liying.liu, zuoguan.wang\}@bst.ai}
  } \\[0.2em]
  {\normalfont\small
    $^{*}$Equal contribution \quad
    $^{\dagger}$Technical lead \quad
    $^{\ddagger}$Corresponding author
  } \\
  {\normalfont\small
    $^{\S}$Work done during an internship at Black Sesame Technologies
  }
}

\newcommand{\projecturl}{\url{https://youtu.be/xVRq1oDzBrw?si=WRPDBOJsYRoR74FE}}

\begin{document}
\maketitle

\begin{abstract}
Recent perception-free end-to-end (E2E) autonomous driving methods bypass explicit perception outputs by compressing dense image patch tokens into compact scene tokens for downstream trajectory generation and scoring.
While these scene tokens form a compact visual bottleneck for the planner, they receive supervision solely from the planning objective, providing limited constraints on the encoded visual information.
To address this limitation, we introduce Neural Token Reconstruction (NTR), a representation learning framework to directly constrain the compact scene-token bottleneck in perception-free driving.
NTR introduces a self-distillation masked latent reconstruction objective that reconstructs masked patch-level latent features using only compact scene tokens as reconstruction memory.
This forces reconstruction gradients to pass exclusively through the scene-token bottleneck, encouraging scene tokens to preserve richer and less redundant visual representations for planning.
We further introduce semantic priors derived from foundation-model annotations as a weak semantic interface biasing reconstruction targets toward driving-related structures without introducing explicit perception heads.
All auxiliary reconstruction components are removed at inference time, leaving the deployed planner unchanged.
NTR achieves state-of-the-art performance on three public autonomous driving benchmarks, including \textbf{8.0461 RFS on Waymo E2E and 94.1 PDMS / 90.9 EPDMS on NavSim1\&2}. The learned scene tokens exhibit lower pairwise redundancy and higher effective rank, indicating that effective bottleneck supervision improves both compact visual representation learning and planning performance. We further validate NTR on a private large-scale driving dataset and integrate it into a real-vehicle planning stack, demonstrating its practical applicability beyond public benchmarks. Real-world deployment videos are available at \projecturl.
\end{abstract}


\section{Introduction}
Perception-free end-to-end (E2E) autonomous driving 
methods~\cite{bojarski2016end,codevilla2019exploring,wang2026drive,guo2025ipaditerativeproposalcentricendtoend,feng2025rap} 
map image observations directly to future trajectories, without 
producing explicit perception outputs such as detection, segmentation, 
and occupancy~\cite{hu2023uniad,jiang2023vad,jiang2026vadv,li2025wote,sun2025sparsedrive,sun2026sparsedrivev2,10670964}. 
In this setting, the learned visual representation alone 
determines what scene information reaches the planner. 
Previous SOTA perception-free methods~\cite{kirby2026drivor,ang2026clover} adopt 
the token-compression idea from image 
tokenizers~\cite{yu2024an,bachmann2025flextok}, reducing dense 
Vision Transformer (ViT) patch features into a compact set of 
scene tokens. These scene tokens serve as the only visual interface 
between visual backbone and planner, forming a many-to-few compression 
bottleneck.

Previous SOTA methods~\cite{kirby2026drivor,ang2026clover} supervise this bottleneck solely through trajectory-level planning objectives, which provide only weak constraints on the fine-grained visual information preserved during compression. As a result, scene tokens tend to learn redundant representations with highly overlapping spatial attention patterns, as illustrated in Fig.~\ref{fig:cover}(a). This failure mode is closely related to a known issue in Vision Transformers, where compact aggregation tokens are processed identically to patch tokens despite their fundamentally different roles~\cite{marouani2026revisiting}. Without explicit supervision, such tokens often drift toward lazy aggregation and redundant encoding~\cite{shi2026visiontransformersneedregisters}.  Existing efforts primarily focus on architectural choices such as the number or structure of scene tokens~\cite{kirby2026drivor,bachmann2025flextok,yu2024an}, but largely overlook the representation quality of the compact planning bottleneck itself. This raises a fundamental representation learning question in perception-free planning:
\begin{figure}[!t]
    \centering
    \begin{tabular}{c@{\hspace{0.02\textwidth}}c}
        \includegraphics[width=0.47\linewidth]{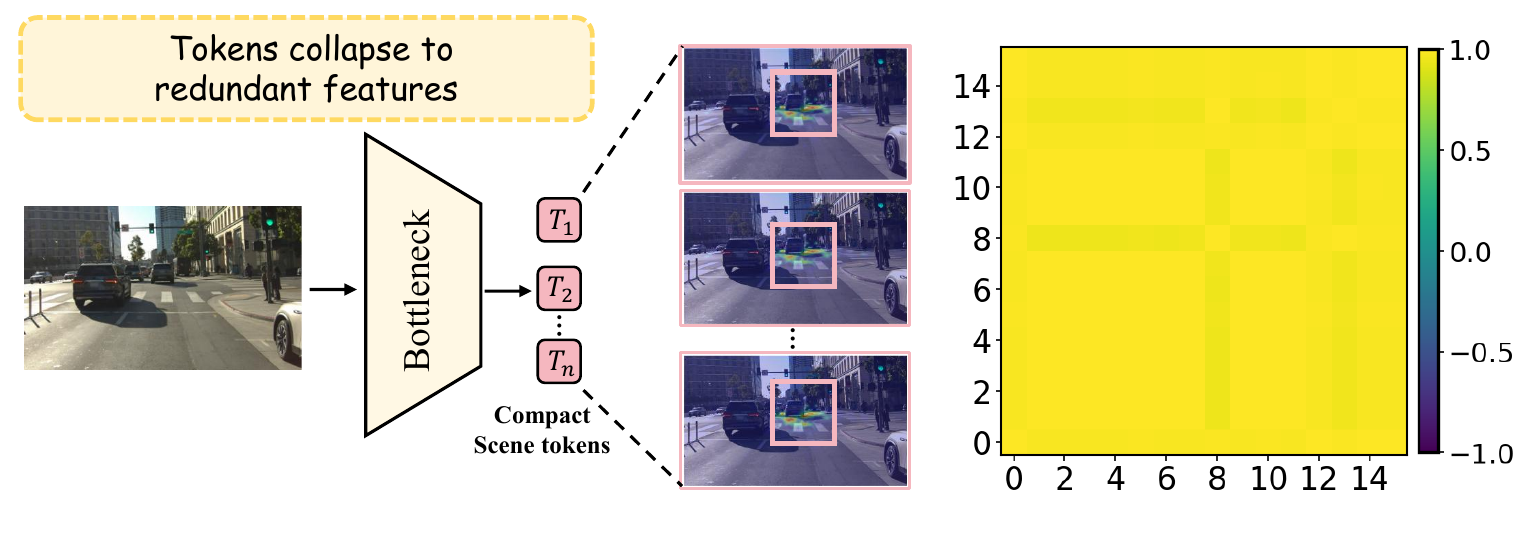} &
        \includegraphics[width=0.47\linewidth]{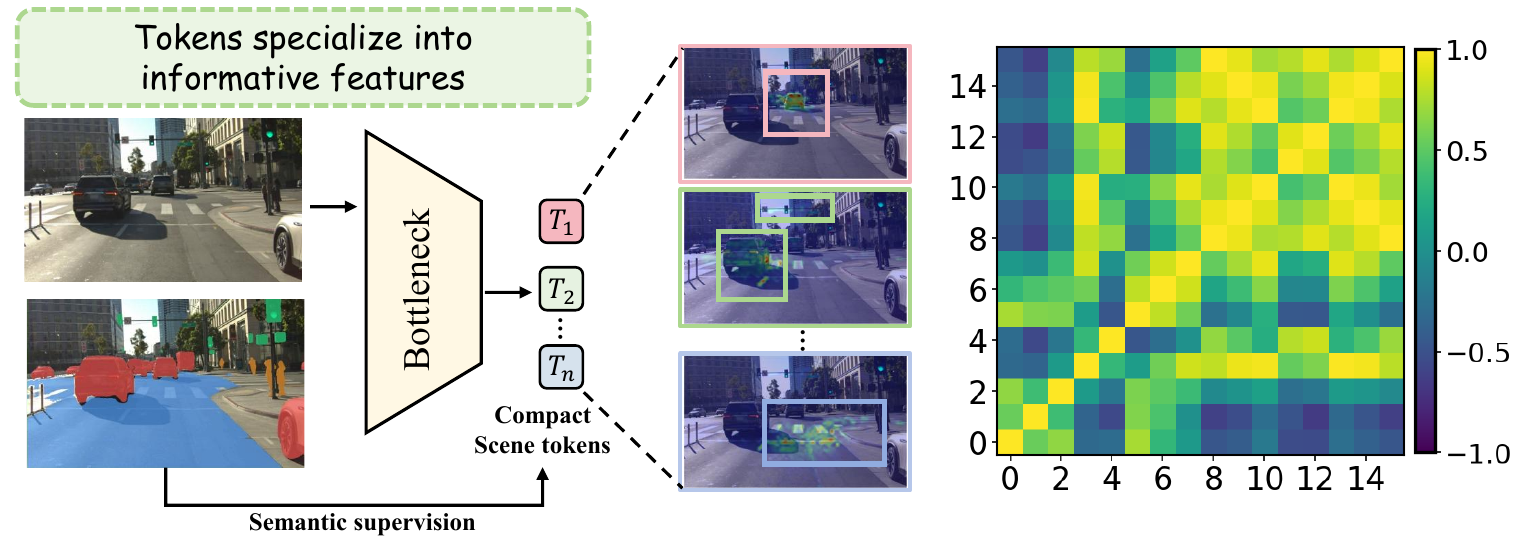} \\
        (a) Redundant token representations & (b) Informative token representations
    \end{tabular}
    \vspace{-0.1cm}
\caption{
Diagnostic visualization of scene-token behavior.
(a) The planning-only baseline shows overlapping attended regions and high token similarity.
(b) NTR shows broader coverage over structured scene elements and lower token similarity.
}
    \label{fig:cover}
    \vspace{-1.5em}
\end{figure}

\begin{quote}
\it
How can compact scene tokens preserve planning-relevant visual information without explicit perception supervision?
\end{quote}

To answer this question, we propose Neural Token Reconstruction (NTR), a self-distillation framework that directly supervises the scene-token bottleneck through masked latent reconstruction. Instead of reconstructing images, NTR reconstructs patch-level latent representations using only compact scene tokens as memory, explicitly encouraging the bottleneck to preserve informative and complementary local visual evidence for planning. During training, an online ViT encoder~\cite{Oquab2023DINOv2,simeoni2025dinov3} processes masked images to generate scene tokens, while an EMA teacher~\cite{assran2023selfsupervisedlearningimagesjointembedding,assran2025vjepa2selfsupervisedvideo} operating on clean images provides stop-gradient latent targets. A lightweight reconstruction decoder is attached only during training, and reconstruction gradients reach the encoder exclusively through the compact scene-token bottleneck. To further improve reconstruction quality, we introduce semantic priors automatically derived from foundation-model segmentation to focus supervision on structured and safety-critical driving regions, including agents, drivable areas and traffic-control elements.
Importantly, the reconstruction branch, EMA teacher, and semantic priors are used only during training and introduce zero additional inference-time computation or architectural changes to the planner.

Extensive experiments on Waymo Open Dataset and NavSim V1\&2 demonstrate that NTR consistently improves planning performance across multiple public autonomous driving benchmarks. Beyond benchmark gains, NTR also reduces scene-token redundancy and improves representation diversity, suggesting that direct bottleneck supervision leads to more informative compact planning representations. We further validate NTR on a large-scale private driving dataset and integrate it into a real-vehicle planning stack, demonstrating strong scalability and practical applicability in real-world autonomous driving systems. The key contribution can be summarized as:
\begin{itemize}
    \item We identify the scene-token bottleneck as a core representation learning challenge in perception-free end-to-end autonomous driving, and propose Neural Token Reconstruction (NTR), a framework that directly supervises compact planning representations through masked latent reconstruction.

    \item We introduce semantic-guided reconstruction priors that focus supervision on structured and safety-critical driving regions without introducing explicit perception heads or inference-time overhead.
    
    \item Extensive experiments on Waymo E2E and NavSim V1\&2 demonstrate that NTR significantly improves planning performance while reducing scene-token redundancy and improving representation diversity. Additional large-scale real-world experiments and vehicle deployment validate its scalability and practical applicability.
\end{itemize}


\section{Related Work}
\label{sec:citations}

\textbf{End-to-End Autonomous Driving.}
End-to-end autonomous driving methods can generally be divided into perception-based and perception-free paradigms. Perception-based methods supervise the visual representation 
through auxiliary perception tasks such as detection, tracking, map 
segmentation, and occupancy 
prediction~\cite{hu2023uniad,wozniak2025prix,sun2025sparsedrive,chitta2023transfuser,li2022bevformer}, 
which require dense annotations and task-specific sub-modules. 
Perception-free methods drop these auxiliary perception tasks and 
feed the features produced by an image backbone directly to the 
planner. Early work mapped sensor input to control commands through 
behavior cloning~\cite{bojarski2016end,codevilla2019exploring}. 
Recent perception-free planners compress dense patch features into 
a compact representation before planning through register-style token 
compression~\cite{kirby2026drivor,ang2026clover} inspired by compact tokenizers~\cite{yu2024an,bachmann2025flextok}. However, the compression is supervised only by the planning loss, which 
provides a weak signal for what visual evidence the compact 
representation should retain. Our work builds on register-style perception-free planners and studies the compact scene-token interface explicitly as a representation bottleneck for planning.

\textbf{Masked Representation Learning in Driving.}
Masked prediction and self-distillation are standard paradigms 
for self-supervised visual representation learning, in both 
pixel~\cite{He2022MAE,Tong2022VideoMAE} and latent 
space~\cite{Oquab2023DINOv2,assran2023selfsupervisedlearningimagesjointembedding,assran2025vjepa2selfsupervisedvideo,baevski2022data2vecgeneralframeworkselfsupervised,zhou2022image}. 
In autonomous driving, related objectives have been used to 
predict future scene latents within world 
models~\cite{li2026drivevlaw0,chen2025drivinggpt} or to 
pretrain ViT backbones for planning relevance through 
JEPA-style self-supervision~\cite{wang2026drive}. Both 
directions operate on the visual backbone or its dense 
feature map and require a separate pretraining stage before 
the planning objective is introduced. NTR applies masked 
latent reconstruction at the compact scene-token bottleneck 
instead. The reconstruction decoder cross-attends only to 
scene tokens, so the auxiliary signal constrains what 
information is retained after compression and is optimized 
jointly with the planning losses in a single end-to-end stage.

\textbf{Foundation Models for Driving and Robotic Perception.}
Foundation models are increasingly used as a source of 
perception information or as a substitute for explicit perception 
modules in driving and robotics. 
Recent works use pretrained geometric or perception foundation models to provide spatial representations~\cite{Wang_2026_CVPR,zhang2026from}, explicit perception outputs~\cite{zeng2025futuresightdrive}, or reconstruction targets for downstream policy learning~\cite{song2025reconvla}.
Our work uses foundation-model segmentation differently.
Instead of supplying explicit perception outputs to the planner, we use SAM3-generated masks~\cite{carion2026sam} only as weak semantic priors for reconstruction target allocation.
These priors bias reconstruction toward structured driving regions such as agents, drivable areas, and traffic-control elements, while avoiding reconstruction budget on uninformative background regions.
All semantic priors are removed during inference and do not introduce additional perception modules or deployment overhead.

\section{Methodology}

\begin{figure}[htbp]
    \centering
    \begin{overpic}[width=\linewidth]{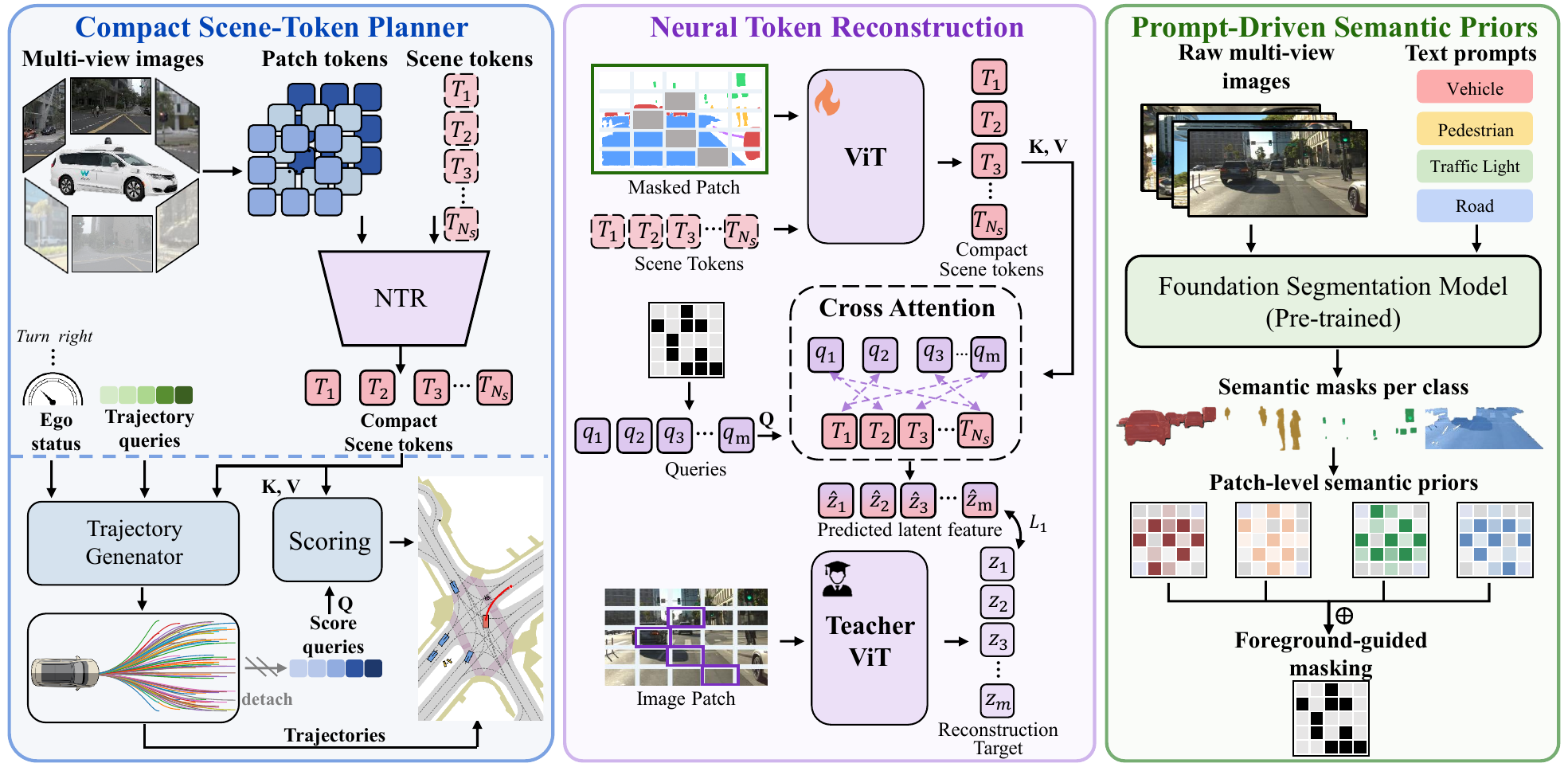}
    \end{overpic}
    \vspace{-5mm}
    \caption{
    Overview of Neural Token Reconstruction (NTR).
    \textcolor[HTML]{0345B7}{\textbf{(a) Compact Scene-Token Planner:} } The compact scene token planner compresses patch tokens into queryable scene tokens for trajectory generation and scoring.
    \textcolor[HTML]{7A2EC0}{\textbf{(b) Neural Token Reconstruction:} } NTR supervises this bottleneck by reconstructing masked teacher features using only the scene tokens as memory.
    \textcolor[HTML]{246C0B}{\textbf{(c) Prompt-Driven Semantic Priors:} } Semantic priors from foundation annotations allocate reconstruction targets toward structured driving elements.
    }
    \label{fig:overview}
    \vspace{-4mm}
\end{figure}

Figure~\ref{fig:overview} gives an overview of NTR.
Our framework keeps the planner architecture unchanged and adds supervision only at training time.

\vspace{-2mm}

\subsection{Compact Scene Token Bottleneck}
\label{sec:preliminaries}

We adopt a DrivoR-style planner~\cite{kirby2026drivor} as the base architecture, and use it to study how compact scene tokens retain visual information for planning. We describe the method for a single planning frame and omit the frame index throughout this section. At current timestep, the input consists of $V$ synchronized camera images $\mathcal{I}=\{I^v\}_{v=1}^{V}$ and the current ego state $e=[v,a,\psi,c]$, where $v$, $a$, $\psi$, and $c$ denote ego velocity, acceleration, heading, and high-level navigation command. Each view image is tokenized into a patch grid of size $H_p \times W_p$, producing patch tokens
$\mathbf{P}^v \in \mathbb{R}^{H_pW_p \times D}$ for view $v$, where $D$ denotes the token feature dimension. The base architecture introduces $N_s$ learnable scene tokens $\mathbf{S}^v \in \mathbb{R}^{N_s \times D}$, where $N_s \ll H_pW_p$. These scene tokens interact with patch tokens from the same view inside the ViT backbone $f_{\theta}$, where LoRA~\cite{hu2021loralowrankadaptationlarge} adapters are used for efficient fine-tuning:
\begin{equation}
\tilde{\mathbf{S}}^v,\tilde{\mathbf{P}}^v
=
f_{\theta}(\mathbf{S}^v,\mathbf{P}^v),
\qquad v=1,\dots,V.
\end{equation}
where only $\tilde{\mathbf{S}}^v$ is exposed to the downstream planner, while patch features are discarded. 

We write the per-view compression from dense patch tokens to scene tokens as
\begin{equation}
\mathbf{P}^v
\in \mathbb{R}^{H_p W_p \times D}
\quad \longrightarrow \quad
\tilde{\mathbf{S}}^v \in \mathbb{R}^{N_s \times D},
\qquad
N_s \ll H_pW_p .
\end{equation}
This mapping forms the scene token bottleneck between image features and planning.
After compression, the downstream planner no longer accesses the $H_pW_p$ dense patch tokens from each view and relies on the corresponding $N_s$ scene tokens.

Finally, we concatenate scene tokens from all views as
$\tilde{\mathbf{S}}=\mathrm{Concat}_{v=1}^{V}(\tilde{\mathbf{S}}^v)$.
The resulting $\tilde{\mathbf{S}}$ is the only visual key-value memory exposed to the downstream planner.
They therefore determine what image evidence remains available for trajectory generation and scoring. 
We abstract the downstream trajectory generation and scoring module as
\begin{equation}
\{(\hat{\tau}_{k},\hat{s}_{k})\}_{k=1}^{N_q}
=
\mathcal{D}_{\phi}(\tilde{\mathbf{S}},e),
\end{equation}
where $\hat{\tau}_{k}$ and $\hat{s}_{k}$ denote the $k$-th candidate trajectory and predicted score. 
In our implementation, $\mathcal{D}_{\phi}$ follows the DrivoR-style trajectory proposal and trajectory-conditioned scoring design. Candidate trajectories are generated from the scene token memory and ego state, and each candidate is scored using the same scene memory. 
NTR supervises the construction of $\tilde{\mathbf{S}}$ and leaves $\mathcal{D}_{\phi}$ unchanged. 
Additional architectural details are provided in Appendix~\ref{app:planner}. 

\subsection{Neural Token Reconstruction}
\label{sec:ntr}

The scene token bottleneck defined in Section~\ref{sec:preliminaries} receives no explicit supervision in the base planner.
It is optimized only indirectly through trajectory regression and candidate scoring.
NTR adds a direct and dense training signal to this bottleneck by reconstructing masked teacher features from the scene tokens used by the planner.
The reconstruction branch is used only during training and does not modify the trajectory generation or scoring modules at inference time.

NTR follows a self-distillation design with an online branch and an EMA teacher branch.
During training, the online encoder receives masked patch tokens together with the learnable scene tokens:
\begin{equation}
\tilde{\mathbf{S}}^{v,m}, \tilde{\mathbf{P}}^{v,m}
=
f_{\theta}(\mathbf{S}^{v}, \mathbf{P}^{v,m}) .
\end{equation}
We keep $\tilde{\mathbf{S}}^{v,m}$ as the scene token memory for both downstream planning and reconstruction.
The dense online patch outputs $\tilde{\mathbf{P}}^{v,m}$ are not used as reconstruction memory.

The teacher branch processes clean patch tokens and provides latent targets.
The teacher encoder $f_{\xi}$ is an EMA copy of the online encoder and is evaluated without gradient updates:
\begin{equation}
\tilde{\mathbf{S}}^{v,*}, \bar{\mathbf{Z}}^{v}
=
f_{\xi}(\mathbf{S}^{v}, \mathbf{P}^{v}),
\qquad
\mathbf{Z}^{v,*}
=
\mathrm{sg}(\bar{\mathbf{Z}}^{v}) .
\end{equation}
Here, $\mathrm{sg}(\cdot)$ denotes stop-gradient, and only the teacher patch outputs $\mathbf{Z}^{v,*}$ are used as reconstruction targets.
The teacher parameters are updated by EMA, $\xi \leftarrow \mu \xi + (1-\mu)\theta$, where $\mu$ is the momentum coefficient.

Let $\rho_{\mathrm{rec}}\in(0,1)$ denote the reconstruction ratio.
For each view $v$, we sample a reconstruction location set $\Omega_{\mathrm{rec}}^{v}\subset\{1,\dots,H_pW_p\}$ with $|\Omega_{\mathrm{rec}}^{v}|=\lfloor\rho_{\mathrm{rec}}H_pW_p\rfloor$.
The patch tokens at these locations are masked in the online branch and used as prediction targets.
The set $\Omega_{\mathrm{rec}}^{v}$ is sampled either uniformly or according to the semantic priors described in Section~\ref{sec:priors}.
The selected locations initialize reconstruction queries from patch positional encodings,
$\mathbf{Q}_{\Omega_{\mathrm{rec}}^v}^{\mathrm{pos}}
=
\mathbf{1}_{|\Omega_{\mathrm{rec}}^v|}\mathbf{q}_{\mathrm{rec}}^{\top}
+
\mathbf{E}_{\mathrm{pos}}[\Omega_{\mathrm{rec}}^v]$,
where $\mathbf{q}_{\mathrm{rec}}\in\mathbb{R}^{D}$ is a shared learnable reconstruction query and
$\mathbf{E}_{\mathrm{pos}}\in\mathbb{R}^{H_pW_p\times D}$ is the patch positional encoding.

A lightweight Transformer decoder $g_{\omega}$ predicts teacher features at the selected locations.
The decoder takes positional reconstruction queries as input and cross-attends only to the masked-view scene tokens:
\begin{equation}
\hat{\mathbf{Z}}_{\Omega_{\mathrm{rec}}^v}^{v}
=
g_{\omega}
\left(
\mathbf{Q}_{\Omega_{\mathrm{rec}}^v}^{\mathrm{pos}},
\tilde{\mathbf{S}}^{v,m}
\right).
\end{equation}
Since $g_{\omega}$ has no access to dense online patch outputs, the reconstruction gradient reaches the online encoder through the scene tokens:
\begin{equation}
\frac{\partial \mathcal{L}_{\mathrm{rec}}}{\partial \theta}
=
\sum_{v}
\frac{\partial \mathcal{L}_{\mathrm{rec}}}{\partial \tilde{\mathbf{S}}^{v,m}}
\frac{\partial \tilde{\mathbf{S}}^{v,m}}{\partial \theta}.
\end{equation}
This makes masked latent reconstruction a dense training signal on the bottleneck, complementing the sparse trajectory and score signals from the planning objective. The detailed decoder architecture is provided in Appendix~\ref{app:ntr_decoder}.

\subsection{Reconstruction Allocation with Semantic Priors}
\label{sec:priors}

Uniform reconstruction provides a general dense signal to the scene-token bottleneck, but it does not distinguish structured scene evidence from low-level appearance patterns.
In driving scenes, large image regions may correspond to sky, background texture, or distant clutter, while agents, drivable areas and traffic cues occupy only a subset of patches.
We therefore use semantic priors as a soft allocation signal that biases reconstruction target selection toward these structured driving elements.

Given a prompt set $\mathcal{P}=\{p_k\}_{k=1}^{K}$ covering driving-related categories, a foundation segmentation model produces image-level masks for each camera view.
We denote the generated masks as $\mathbf{U}\in\{0,1\}^{V\times K\times H\times W}$ and pool them to the planner patch grid as $\mathbf{G}=\mathrm{Pool}(\mathbf{U})\in\mathbb{R}^{V\times K\times H_p\times W_p}$.
After normalization along the semantic dimension, $\mathbf{G}(v,k,i,j)$ gives the prior score of category $k$ at patch location $(i,j)$ in view $v$.
We convert the semantic priors into a reconstruction score map:
\begin{equation}
\mathbf{A}(v,i,j)
=
\sum_{k=1}^{K}
\alpha_k\,\mathbf{G}(v,k,i,j)
+
\tau\,\varepsilon_{vij},
\qquad
\varepsilon_{vij}\sim\mathcal{N}(0,1),
\end{equation}
where $\alpha_k$ is the category weight and $\tau$ controls stochastic exploration.
The selected reconstruction locations are then obtained by taking the top-scoring patches over the full patch grid under the same reconstruction ratio $\rho_{\mathrm{rec}}$ used in Section~\ref{sec:ntr}:
\begin{equation}
\Omega_{\mathrm{rec}}^{v}
=
\mathrm{Top}\text{-}m
\left(
\mathbf{A}(v,:,:)
\right),
\qquad
m=
\left\lfloor
\rho_{\mathrm{rec}}H_pW_p
\right\rfloor .
\end{equation}
Here $\Omega_{\mathrm{rec}}^{v}\subset\{1,\dots,H_pW_p\}$ denotes the selected patch index set after flattening the patch grid.
The selected locations $\Omega_{\mathrm{rec}}^{v}$ are used both to mask the online input patches and to define the support of the reconstruction loss.
For the random masking baseline, we replace $\mathbf{A}$ with random patch scores while keeping the same reconstruction ratio.
The semantic priors are used only during training to allocate reconstruction targets, and are removed together with the reconstruction branch at inference time.
Detailed prompt-based semantic prior generation pipeline is illustrated in Appendix~\ref{app:semantic_prior}.

\subsection{Training Objectives}
\label{sec:training}

The overall objective combines the original planning loss with the NTR reconstruction loss:
\begin{equation}
\mathcal{L}
=
\mathcal{L}_{\mathrm{plan}}
+
\lambda_{\mathrm{rec}}
\mathcal{L}_{\mathrm{rec}},
\end{equation}
where $\lambda_{\mathrm{rec}}$ controls the strength of the reconstruction objective. 
$\mathcal{L}_{\mathrm{plan}}$ follows the base planner and includes trajectory supervision and score supervision for trajectory selection. 
Detailed planning losses for different benchmarks are provided in Appendix~\ref{app:planning loss}. 

We collect the reconstruction locations from all camera views as
\begin{equation}
\Omega_{\mathrm{rec}}
=
\{(v,i)\mid i\in\Omega_{\mathrm{rec}}^{v},\ v=1,\dots,V\}.
\end{equation}
The reconstruction loss is computed in the teacher latent feature space:
\begin{equation}
\mathcal{L}_{\mathrm{rec}}
=
\frac{1}{|\Omega_{\mathrm{rec}}|D}
\sum_{(v,i)\in\Omega_{\mathrm{rec}}}
\left\|
\hat{\mathbf{z}}_{i}^{v}
-
\mathbf{z}_{i}^{v,*}
\right\|_1 .
\end{equation}
Here, $i$ indexes a patch location in view $v$, $D$ denotes the hidden dimension, $\hat{\mathbf{z}}_{i}^{v}$ is the reconstructed latent feature, and $\mathbf{z}_{i}^{v,*}$ is the corresponding stop-gradient teacher target. 
Semantic priors only determine the support of $\Omega_{\mathrm{rec}}^{v}$.


\section{Experiments}


\subsection{Experimental Setup}

\paragraph{Datasets and Evaluation Metrics.}
We evaluate NTR on the Waymo E2E dataset~\cite{xu2025wode2ewaymoopendataset} and NavSim V1\&2~\cite{dauner2024navsim,cao2025navsimv2}.
Waymo is evaluated with Rater Feedback Score (RFS), which is based on human ratings, while NavSim uses closed-loop proxy scores based on driving rules, including PDMS and its extension EPDMS.
On Waymo, we train on 2,037 segments, corresponding to approximately 76k samples, and evaluate on 479 validation segments and 1,505 test segments.
On NavSim, we use the full navtrain split with approximately 103k samples.

\textbf{Implementation Details.}
We use DrivoR~\cite{kirby2026drivor} as the baseline planner, with a pretrained DINOv2 ViT-S backbone~\cite{Oquab2023DINOv2} adapted by LoRA~\cite{hu2021loralowrankadaptationlarge}.
Unless otherwise specified, we use four camera views and 16 scene tokens per camera.
The reconstruction ratio is set to $\rho_{\mathrm{rec}}=0.3$, the EMA momentum is set to $\mu=0.999$, and the temperature is set to $\tau=0.4$.
The reconstruction loss weight $\lambda_{\mathrm{rec}}$ is set to 1.0 on Waymo and 0.5 on NavSim.
We use the AdamW optimizer~\cite{loshchilov2019decoupledweightdecayregularization} and train on a cluster of 8 NVIDIA A800 GPUs with a total batch size of 128.
Semantic priors are generated once offline with SAM3 and cached.
The learning rate is initialized to $2\times10^{-4}$ and scheduled with cosine annealing after a warmup over the first 10\% of training steps.
Full benchmark protocols, optimization hyperparameters, and implementation details are provided in Appendix~\ref{app:benchmark_training}.

\subsection{Experiment Results}
\label{sec:main_results}

\textbf{Benchmark Results.} Table~\ref{tab:waymo_short} reports the results on the Vision-based End-to-End Driving track of the Waymo Open Dataset Challenge. 
This table compares our final NTR submission with existing leaderboard methods. 
NTR achieves the best overall RFS among the listed methods, reaching 7.9982 with a single model and 8.0461 with ensembling. 
It also obtains the best ADE@5s and ADE@3s, showing that the improvement is reflected in both the human-rating-based score and trajectory accuracy. 
To isolate the net gain of NTR, we also submit a matched DrivoR baseline trained for 30 epochs with the same data, backbone, token budget, planner and optimization schedule, so that NTR is the only training-time addition.
NTR improves this baseline by 3.68\% in RFS (7.7146 $\rightarrow$ 7.9982) and reduces ADE@5s and ADE@3s by 10.30\% and 7.95\%, respectively.
Our ensemble combines two identical NTR models that differ only in the random seed (single-model RFS 7.9982 and 7.9445), using RAP-style candidate pooling~\cite{feng2025rap}, at roughly $2\times$ the inference cost (Appendix~\ref{app:ensemble}).
Table~\ref{tab:navsim_main} and Table~\ref{tab:navsim_v2} further evaluate NTR on navtest with PDMS and EPDMS. 
Using only the standard navtrain split without extra simulated data, NTR achieves 94.1 PDMS and 90.9 EPDMS, providing complementary evidence under a closed-loop planning metric. 

\begin{table*}[!htbp]
\centering
\caption{Performance on the leaderboard of the Vision-based End-to-End Driving track of the Waymo Open Dataset Challenge. $^{\dagger}$Trained by us with the same data, backbone, token budget, planner and 30-epoch schedule as NTR. Bold indicates best.}
\label{tab:waymo_short}
\setlength{\tabcolsep}{4pt}
\resizebox{\linewidth}{!}{%
\begin{tabular}{ll|ccccccc}
\hline
Method & Reference & RFS $\uparrow$ & ADE 5s $\downarrow$ & ADE 3s $\downarrow$ & RFS (Cons.) $\uparrow$ & RFS (Inter.) $\uparrow$ & RFS (Ped.) $\uparrow$ & RFS (Cyc.) $\uparrow$ \\ \hline

RAP (Ensemble) & ICLR'26\cite{feng2025rap} & 8.0430 & 2.6457 & 1.1741 & 8.6939 & \textbf{8.1798} & 8.0336 & 7.7604 \\
Poutine & arXiv'25\cite{rowe2025poutine} & 7.9860 & 2.7419 & 1.2055 & 8.3595 & 8.1356 & 7.9529 & 7.8325 \\
IRL-VLA & arXiv'25\cite{jiang2025irlvla} & 7.8900 & 2.8235 & 1.2184 & 8.4401 & 7.9781 & 7.9236 & 7.7757 \\
FROST-Drive & arXiv'26\cite{dong2026frostdrive} & 7.8560 & 3.5653 & 2.5373 & 8.2510 & 7.9658 & 7.9293 & 7.7192 \\
HMVLM & arXiv'25\cite{wang2025hmvlm} & 7.7367 & 3.0715 & 1.3269 & 8.6663 & 7.9043 & 7.8578 & 7.3925 \\
dVLM-AD & arXiv'25\cite{ma2025dvlmad}  & 7.6331 & 3.0221 & 1.2849 & 8.1177 & 7.9692 & 7.6775 & 7.5778 \\
AutoVLA & NeurIPS'25\cite{zhou2025autovla}  & 7.5566 & 2.9580 & 1.3507 & 7.9556 & 7.7112 & 7.5920 & 7.3208 \\
Waymo Baseline & arXiv'25\cite{xu2025wode2e}  & 7.5281 & 3.0182 & 1.3200 & 8.2729 & 7.6226 & 7.6651 & 7.5172 \\
DrivePI & arXiv'25\cite{liu2025drivepi} & 7.5160 & 2.9066 & 1.2526 & 8.2487 & 7.8302 & 7.9286 & 7.7029 \\

OpenEMMA w/ Qwen & arXiv'25\cite{xing2025openemma} & 6.3162 & 4.1078 & 1.8663 & 6.3312 & 6.7374 & 6.6844 & 5.9696 \\
\hline
DrivoR$^{\dagger}$ (matched baseline) & CVPR'26\cite{kirby2026drivor} & 7.7146 & 3.0715 & 1.2928 & 8.4069 & 8.0146 & \textbf{8.1016} & 7.6846 \\
\rowcolor[rgb]{0.9,0.9,0.9}
NTR (Single) & Ours & 7.9982 & 2.7550 & 1.1900 & \textbf{8.9490} & 8.1064 & 7.9781 & 7.8125 \\
\rowcolor[rgb]{0.9,0.9,0.9}
NTR (Ensemble) & Ours & \textbf{8.0461} & \textbf{2.6379} & \textbf{1.1729} & 8.8159 & 8.1237 & 8.0764 & \textbf{7.9158} \\ \hline

\end{tabular}%
}
\vspace{-2em}
\end{table*}

\begin{table}[!htbp]
\centering
\caption{Performance on navtest split evaluated with NavSim v1 metrics. Bold indicates best.}
\label{tab:navsim_main}
\setlength{\tabcolsep}{5pt}
\resizebox{\linewidth}{!}{%
\begin{tabular}{ll|ccccccc}
\hline
Method & Reference & NC $\uparrow$ & DAC $\uparrow$ & TTC $\uparrow$ & Comf. $\uparrow$ & EP $\uparrow$ & PDMS $\uparrow$ \\ 
\hline
Human Driver(Ground Truth) & NeurIPS'24\cite{dauner2024navsim}  & 100 & 100 & 100 & 99.9 & 87.5 & 94.8 \\
PDM-Closed & PMLR'23\cite{dauner2023parting}   & 94.6 & 99.8 & 89.9 & 86.9 & 99.9 & 89.1 \\
\hline
AutoVLA       & NeurIPS'25\cite{zhou2025autovla}       & 98.4 & 95.6 & \textbf{98.0} & 99.9 & 81.9 & 89.1 \\
DriveVLA-W0   & ICLR'26\cite{li2026drivevlaw0}       & 98.7 & \textbf{99.1} & 95.3 & 99.3 & 83.3 & 90.2 \\
RAP             & ICLR'26\cite{feng2025rap}               & \textbf{99.1} & 98.9 & 96.7 & \textbf{100.0} & 90.3 & 93.8 \\
TransFuser      & TPAMI'22\cite{chitta2022transfuser}    & 97.7 & 92.8 & 92.8 & \textbf{100.0} & 79.2 & 84.0 \\
DiffusionDrive  & CVPR'25\cite{liao2025diffusiondrive}    & 98.2 & 96.2 & 94.7 & \textbf{100.0} & 82.2 & 88.1 \\
GoalFlow        & CVPR'25\cite{xing2025goalflow}        & 98.4 & 98.3 & 94.6 & \textbf{100.0} & 85.0 & 90.3 \\
DrivingGPT      & ICCV'25\cite{chen2025drivinggpt}         & 98.9 & 90.7 & 94.9 & 95.6 & 79.7 & 82.4 \\
UniAD           & CVPR'23\cite{hu2023uniad}               & 97.8 & 91.9 & 92.9 & \textbf{100.0} & 78.8 & 83.4 \\
VAD-v2          & ICLR'26\cite{jiang2026vadv}              & 98.1 & 94.8 & 94.3 & \textbf{100.0} & 80.6 & 86.2 \\
ReCogDrive      & ICLR'26\cite{li2026recogdrive}          & 97.9 & 97.3 & 94.9 & \textbf{100.0} & 87.3 & 90.8 \\
DrivoR          & CVPR'26\cite{kirby2026drivor}          & 98.9 & 98.3 & 96.2 & \textbf{100.0} & 89.1 & 93.7 \\
\hline
\rowcolor[rgb]{0.9,0.9,0.9}
\textbf{NTR} & \textbf{Ours}  & \textbf{99.1} & 98.8 & 96.9 & \textbf{100.0} & \textbf{90.8} & \textbf{94.1} \\
\hline
\end{tabular}%
}
\vspace{-1.5em}
\end{table}

\begin{table*}[!htpb]
\centering
\caption{Performance on navtest split with NavSim v2 extended metrics. Bold indicates best.}
\vspace{-0.5em}
\label{tab:navsim_v2}
\setlength{\tabcolsep}{5pt}
\resizebox{\textwidth}{!}{%
\begin{tabular}{ll|cccccccccc}
\hline
Method & Reference
& NC $\uparrow$ 
& DAC $\uparrow$ 
& DDC $\uparrow$ 
& TL $\uparrow$ 
& EP $\uparrow$ 
& TTC $\uparrow$ 
& LK $\uparrow$ 
& HC $\uparrow$ 
& EC $\uparrow$ 
& EPDMS $\uparrow$ \\
\hline

Ego-MLP 
& CVPR'24\cite{Li_2024_CVPR}
& 93.1 & 77.9 & 92.7 & 99.6 & 86.0 & 91.5 & 89.4 & \textbf{98.3} & 85.4 & 64.0 \\

TransFuser
& TPAMI'22\cite{chitta2022transfuser}
& 96.9 & 89.9 & 97.8 & 99.7 & 87.1 & 95.4 & 92.7 & \textbf{98.3} & 87.2 & 76.7 \\

ARTEMIS
& RAL'26\cite{feng2025artemis}
& 98.3 & 95.1 & 98.6 & 99.8 & 81.5 & 97.4 & 96.5 & \textbf{98.3} & \textbf{98.3} & 83.1 \\

Hydra-MDP++
& arXiv'25\cite{li2025hydramdpplusplus}
& 98.4 & 98.0 & 99.4 & 99.8 & 87.5 & 97.7 & 95.3 & \textbf{98.3} & 77.4 & 85.1 \\

DriveSuprim
& AAAI'25\cite{yao2025drivesuprim}
& 97.8 & 97.9 & 99.5 & \textbf{99.9} & 90.6 & 97.1 & 96.6 & \textbf{98.3} & 77.9 & 86.0 \\

DiffusionDriveV2
& arXiv'25\cite{zou2025diffusiondrivev2}
& 97.7 & 96.6 & 99.2 & 99.8 & 88.9 & 97.2 & 96.0 & 97.8 & 91.0 & 87.5 \\

SparseDriveV2
& ICLR'26\cite{sun2026sparsedrivev2}
& 98.1 & 98.1 & 99.6 & 99.8 & \textbf{91.1} & 97.3 & \textbf{96.9} & 98.2 & 78.4 & 90.1 \\

\hline
\rowcolor[rgb]{0.9,0.9,0.9}
\textbf{NTR}
& \textbf{Ours}
& \textbf{99.4} & \textbf{99.0} & \textbf{99.7} & 99.7 & 88.3 & \textbf{98.9} & 95.5 & 98.0 & 76.9 & \textbf{90.9} \\
\hline

\end{tabular}%
}
\vspace{-1em}
\end{table*}

\begin{figure}[!htbp]
    \centering
    \includegraphics[width=\linewidth]{
    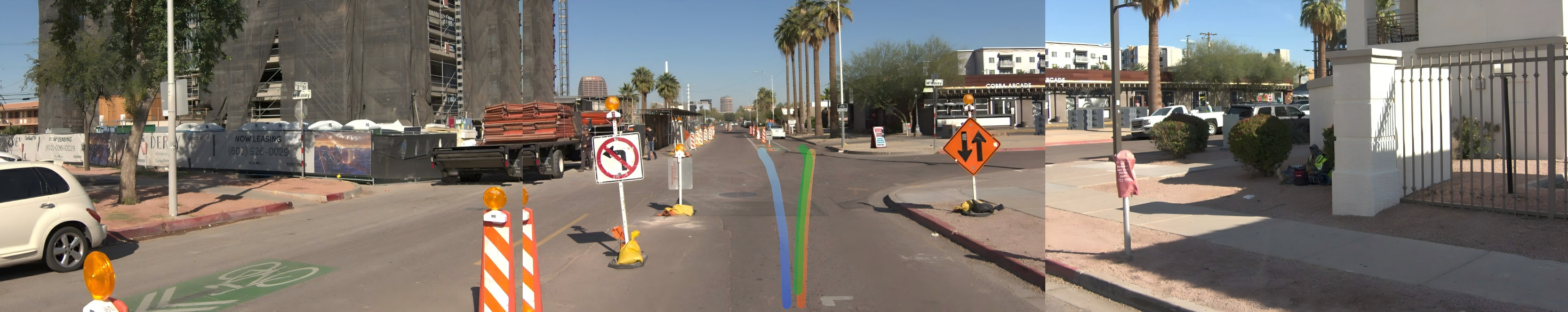}
    \begin{tikzpicture}
        \node[anchor=south west, inner sep=0] (img) at (0,0)
        {\includegraphics[width=\linewidth]{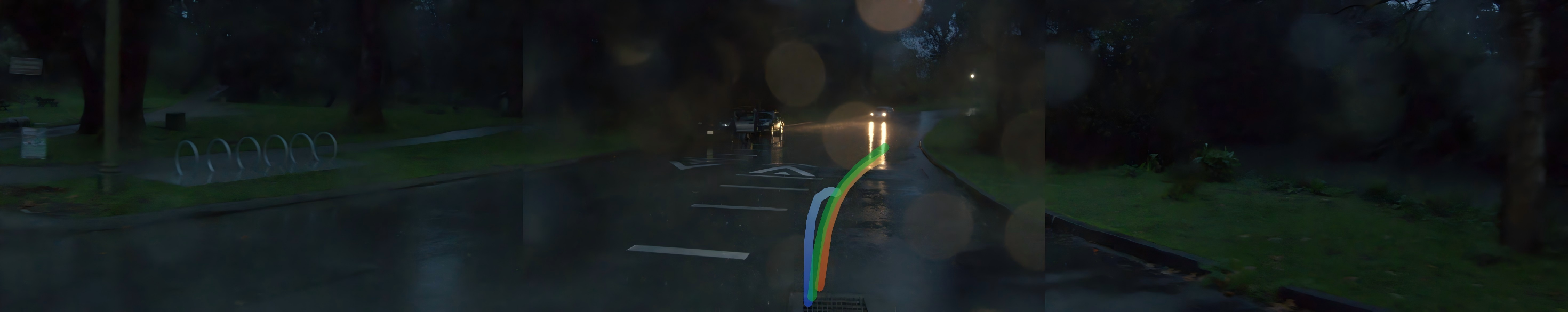}};
        \begin{scope}[x={(img.south east)}, y={(img.north west)}]
            \node[anchor=south east, inner sep=0] at (1,0)
            {\includegraphics[width=0.22\linewidth]{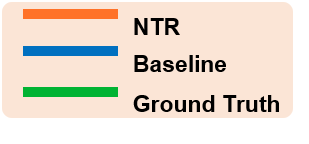}};
        \end{scope}
    \end{tikzpicture}

    \caption{
    Qualitative examples on different driving scenarios.
    }
    \vspace{-2em}
    \label{fig:longtail}
\end{figure}

\textbf{Qualitative Results.}
Fig.~\ref{fig:longtail} provides qualitative planning examples on Waymo E2E.
We compare the final selected trajectories of the planning-only baseline and NTR under different driving scenarios.
In these examples, NTR produces trajectories that are more consistent with the visible road structure and nearby traffic context.
These results provide qualitative evidence for the planning improvement observed in Table~\ref{tab:waymo_short}. 

\textbf{Ablation.}
Table~\ref{tab:core_ablation} isolates the contributions of latent reconstruction, EMA teacher targets, and semantic prior target selection.
Adding masked latent reconstruction with random targets already improves the baseline, indicating that reconstruction provides useful dense supervision for the scene-token bottleneck beyond trajectory and score losses.
Using an EMA teacher further improves RFS from 7.754 to 7.817, suggesting that adaptive but stop-gradient latent targets are more effective than a frozen target encoder.
Semantic prior target selection brings additional gains when combined with both frozen and EMA targets, showing that the location of reconstruction targets also matters.
The full NTR model combines EMA teacher targets with semantic prior allocation and achieves the best performance, improving RFS from 7.652 to 7.974 and reducing ADE@5s from 2.565 to 2.146.

\vspace{-1em}
\begin{table}[!htpb]
\centering
\caption{Core ablation of NTR on the validation split of Waymo E2E Dataset.}
\label{tab:core_ablation}
\resizebox{\columnwidth}{!}{%
\begin{tabular}{lcccccc}
\hline
Method 
& Latent Recon. 
& Target Encoder 
& Recon. Target Selection 
& RFS $\uparrow$ 
& ADE@5s $\downarrow$ 
& ADE@3s $\downarrow$\\
\hline
Baseline 
& None & None & None 
& 7.652 & 2.565 & 1.170\\
Frozen target + Random 
& \checkmark & Frozen & Random 
& 7.754 & 2.255 & 1.024 \\
Frozen target + Semantic Prior
& \checkmark & Frozen & Semantic Prior
& 7.781 & 2.279 & 1.019 \\
EMA target + Random 
& \checkmark & EMA & Random 
& 7.817 & 2.153 & 0.967\\
\rowcolor[rgb]{0.9,0.9,0.9}
NTR 
& \checkmark & EMA & Semantic Prior
& 7.974 & 2.146 & 0.949\\
\hline
\end{tabular}%
}
\end{table}

\textbf{Architectural Generality.}
Table~\ref{tab:generality} evaluates NTR across ViT scales, backbone families and scene-token compressor designs under the same 10-epoch ablation protocol.
NTR consistently improves RFS for learned many-to-few compressors, including the DrivoR-style scene tokens and a Q-Former~\cite{pmlr-v202-li23q} compressor.
Scaling the backbone alone is non-monotonic for the baseline, likely because the scene-token budget and planner capacity are kept fixed.
The gain with parameter-free patch mean-pooling is much smaller, which is consistent with the intended scope of NTR: it targets explicit, learned many-to-few bottlenecks.
Replacing the latent reconstruction target with RGB patches or VAE latents~\cite{Rombach_2022_CVPR} yields 7.739 and 7.866 RFS respectively, below the 7.974 RFS of EMA latent targets (Appendix~\ref{app:alt_targets}).

\begin{table}[!htpb]
\centering
\caption{Generality of NTR across backbone scale, backbone family and compressor design (Waymo E2E validation, RFS).}
\label{tab:generality}
\small
\begin{tabular}{llccc}
\toprule
Axis & Setting & Baseline & +NTR & $\Delta$RFS \\
\midrule
\multirow{2}{*}{Scale}
& DINOv2-S (default) & 7.652 & \textbf{7.974} & +0.322 \\
& DINOv2-B & 7.660 & 7.949 & +0.289 \\
\midrule
Backbone
& DINOv3-S & 7.757 & 7.956 & +0.199 \\
\midrule
\multirow{2}{*}{Compressor}
& Q-Former & 7.737 & 7.930 & +0.193 \\
& Patch mean-pool & 7.865 & 7.882 & +0.017 \\
\bottomrule
\end{tabular}
\vspace{-1em}
\end{table}

\noindent
\begin{minipage}[t]{0.52\linewidth}
\textbf{Token Diagnostics.}
We use two token-wise diagnostics to analyze the compact scene token memory: pairwise cosine similarity and effective rank.
Pairwise cosine similarity measures redundancy among scene token embeddings, while effective rank measures the effective dimensionality of the token feature matrix.
As shown in Table~\ref{tab:token_budget_maintext}, NTR improves RFS across all token budgets.
At the same token budget, NTR consistently reduces token similarity and increases effective rank compared with the planning-only baseline.
These results suggest that NTR makes the compact scene token memory less redundant and better utilized.
A direct token-decorrelation penalty reduces redundancy even further but only reaches 7.695 RFS, showing that diversity alone is insufficient (Appendix~\ref{app:diag_penalty}).
\end{minipage}
\hfill
\begin{minipage}[t]{0.43\linewidth}
\centering
\captionof{table}{
Scene-token budget diagnostics.
}
\label{tab:token_budget_maintext}
\scriptsize
\setlength{\tabcolsep}{2.5pt}
\renewcommand{\arraystretch}{0.92}
\begin{tabular}{@{}l|c|ccc@{}}
\toprule
\textbf{Method} 
& \textbf{\#Tokens}  
& \textbf{RFS}$\uparrow$ 
& \textbf{Eff. Rank}$\uparrow$ 
& \textbf{Sim.}$\downarrow$ \\
\midrule
\multirow{5}{*}{Base.}
& 1  & 7.567& --    & --    \\ 
& 4  & 7.663 & 1.144 & 0.951 \\
& 8  & 7.671 & 1.307 & 0.918 \\
& 16 & 7.652 & 2.080 & 0.953 \\
& 32 & 7.768& 2.133 & 0.850 \\
\midrule
\multirow{5}{*}{NTR}
& 1  & 7.579 & --    & --    \\ 
& 4  & 7.741 & 1.281 & 0.292 \\
& 8  & 7.856 & 2.568 & 0.563 \\
& 16 & 7.974 & 3.380 & 0.407 \\
& 32 & 8.053 & 2.352 & 0.461 \\
\bottomrule
\end{tabular}
\end{minipage}

\textbf{Additional Analyses.}
Semantic-prior sensitivity (prior generator, prompt coverage and category weights), background-attention analysis, additional qualitative results, failure case analysis, real-world deployment details, data scaling experiments and efficiency analysis are provided in the appendix.

\section{Conclusion}
\label{sec:conclusion}

We presented Neural Token Reconstruction (NTR), a training framework that improves perception-free end-to-end planning from a representation learning perspective.
NTR directly supervises the scene-token bottleneck through masked latent reconstruction, encouraging compact scene tokens to retain richer and less redundant visual information before trajectory generation and scoring.
Experiments on Waymo E2E and NavSim show that this training-time supervision improves planning performance while keeping the inference-time planner unchanged.

\section{Limitations}
\label{sec:Limitations}
NTR focuses on supervising the many-to-few compression from dense patch tokens to compact scene tokens, rather than improving the visual backbone itself.
Backbone-level masked modeling methods such as MAE~\cite{He2022MAE,Tong2022VideoMAE}, iBOT~\cite{zhou2022image}, or JEPA~\cite{assran2023selfsupervisedlearningimagesjointembedding,assran2025vjepa2selfsupervisedvideo} are therefore complementary to NTR, as they improve dense patch features before compression but do not directly constrain what information survives in the scene-token memory. A broader comparison with alternative backbone pretraining objectives and token-level regularizers is left for future work.
Its benefit is also tied to learned compressors: with parameter-free patch mean-pooling the gain is marginal.
NTR also relies on foundation-model annotations for semantic-prior allocation, whose quality may vary across domains and rare scenarios, although the results are robust to the choice of prior generator and category weights.
NTR introduces additional preprocessing and training cost (about 2 days of one-time offline prior generation on 8$\times$A800 GPUs and 1.5--2$\times$ training time per epoch), although the deployed planner remains unchanged.
Finally, our real-vehicle experiments serve as deployment validation rather than a statistical safety evaluation, and closed-loop safety under long-tail conditions remains to be established.

\acknowledgments{We thank the anonymous reviewers and the area chair for their constructive feedback, which substantially improved this paper.}


\bibliography{example}  

\clearpage
\appendix
\clearpage
\appendix

\begin{center}
{\Large\bfseries NTR: Neural Token Reconstruction for Scene Token Bottleneck in End-to-End Driving\par}

\vspace{0.5em}
{\large Appendix\par}
\end{center}

\vspace{1em}

This appendix complements the main paper with additional implementation details, benchmark protocols, analyses, and visualizations. It is organized as follows:

\begin{itemize}
    \item Section~\ref{app:benchmark_training} describes benchmark metrics, training configurations, and planning losses.
    \item Section~\ref{app:architecture} details the planner decoder and the NTR reconstruction decoder.
    \item Section~\ref{app:semantic_prior} presents prompt-based semantic prior generation, ablations of the reconstruction allocation, and sensitivity to the prior generator, prompt coverage and category weights.
    \item Section~\ref{app:additional_analysis} provides scene-token diagnostics, including per-token attention analysis, background-attention analysis, a comparison against a direct decorrelation penalty, and alternative reconstruction targets.
    \item Section~\ref{app:efficiency} reports the inference efficiency.
    \item Section~\ref{app:scaling_deployment} includes data scaling and real-world deployment details.
    \item Section~\ref{app:qualitative} presents additional qualitative results and failure cases.
\end{itemize}


  






\section{Benchmark and Training Details}
\label{app:benchmark_training}

\subsection{Benchmark Metrics}
\label{app:benchmark_metrics}
Waymo uses the Rater Feedback Score (RFS), NavSim~v1 uses the Predictive Driver
Model Score (PDMS), and NavSim~v2 uses its extended version (EPDMS). We follow the
official protocols and summarize each below.

\paragraph{Waymo RFS.}
RFS scores a predicted trajectory $\hat{\tau}$ against human rater trajectories
$\{\tau_j\}_{j=1}^{J}$ with preference scores $s_j\in[0,10]$. At each horizon
$T\in\{3\mathrm{s},5\mathrm{s}\}$ the lateral and longitudinal errors to each rater
trajectory are divided by speed-dependent trust-region thresholds (scaled from
$0.5\times$ at low speed up to $1\times$ above $11$\,m/s), giving a normalized
distance $d_j(T)=\max\!\big(|\Delta^{j}_{\mathrm{lat}}|/r_{\mathrm{lat}},\,
|\Delta^{j}_{\mathrm{lng}}|/r_{\mathrm{lng}}\big)$. The prediction takes the
preference score of the closest rater trajectory $j^{*}=\arg\min_j d_j(T)$,
decayed once it leaves the trust region,
\begin{equation}
\mathrm{RFS}_{T}(\hat{\tau})
=
\max\!\big(
s_{\mathrm{floor}},\;
s_{j^{*}}\cdot 0.1^{\max(d_{j^{*}}(T)-1,\,0)}
\big),
\end{equation}
with floor $s_{\mathrm{floor}}=4$ outside the trust region. The final score
averages $\mathrm{RFS}_T$ over horizons, scenarios, and scenario categories;
ADE@3s and ADE@5s are reported as secondary metrics.

\paragraph{NavSim PDMS / EPDMS.}
Both NavSim metrics share one structure, with hard penalty terms entering as a
product and behavior terms as a weighted average,
\begin{equation}
\mathrm{Score}
=
\Big(
\prod_{m\in\mathcal{M}_{\mathrm{pen}}} s_m
\Big)
\cdot
\frac{\sum_{m\in\mathcal{M}_{\mathrm{avg}}} w_m s_m}
     {\sum_{m\in\mathcal{M}_{\mathrm{avg}}} w_m}.
\end{equation}
PDMS (v1) uses $\mathcal{M}_{\mathrm{pen}}=\{\mathrm{NC},\mathrm{DAC}\}$ and
$\mathcal{M}_{\mathrm{avg}}=\{\mathrm{EP},\mathrm{TTC},\mathrm{C}\}$. EPDMS (v2)
extends these to $\mathcal{M}_{\mathrm{pen}}=\{\mathrm{NC},\mathrm{DAC},\mathrm{DDC},
\mathrm{TLC}\}$ and $\mathcal{M}_{\mathrm{avg}}=\{\mathrm{EP},\mathrm{TTC},\mathrm{LK},
\mathrm{HC},\mathrm{EC}\}$ and uses the official filtered sub-scores. The terms are
no-at-fault collision (NC), drivable-area compliance (DAC), driving-direction
compliance (DDC), traffic-light compliance (TLC), ego progress (EP),
time-to-collision (TTC), lane keeping (LK), comfort (C), history comfort (HC), and
extended comfort (EC). EPDMS thus covers a broader set of rule-compliance and
motion-quality constraints than PDMS.

\subsection{Configurations}
\label{app:training_config}
Tables~\ref{tab:training_config} and~\ref{tab:model_config} list the training and
model settings for the main experiments. All ablations train 10 epochs on the Waymo E2E training set, while main results train 30 epochs (training set on Waymo E2E) and 25 epochs (Navtrain split on Navsim). On Waymo we downsample the
trajectories from $10$\,Hz to $2$\,Hz, so the planner predicts $10$ waypoints over
the $5$\,s horizon, and at submission we interpolate the prediction back to the
$4$\,Hz format used by the Rater Feedback Score. The reconstruction branch, EMA
teacher, and semantic priors are training-only and removed at inference, so none
of the ablations change the deployed planner.

\begin{table*}[htbp]
\centering
\caption{Training configurations used in the main experiments. Identical settings across benchmarks are merged for clarity.}
\label{tab:training_config}
\small
\begin{tabular}{p{4.6cm}cc}
\toprule
\textbf{Item} & \textbf{Waymo Open Dataset} & \textbf{NavSim v1} \\
\midrule
GPU & \multicolumn{2}{c}{NVIDIA A800} \\
Num of GPUs & \multicolumn{2}{c}{8} \\
Optimizer & \multicolumn{2}{c}{AdamW} \\
Learning rate & \multicolumn{2}{c}{2e-4} \\
Batch size &  \multicolumn{2}{c}{128} \\
Main training epochs & 30 & 25 \\
Ablation training epochs & 10 & -- \\
EMA teacher & \multicolumn{2}{c}{Enabled for NTR variants} \\
EMA momentum  & \multicolumn{2}{c}{0.999} \\
Reconstruction ratio $\rho_{\mathrm{rec}}$ & \multicolumn{2}{c}{0.3} \\
Noise temperature $\tau$ & \multicolumn{2}{c}{0.4} \\
Reconstruction weight $\lambda_{\mathrm{rec}}$ & 1.0 & 0.5 \\
\bottomrule
\end{tabular}
\end{table*}

\begin{table}[htpb]
\centering
\caption{Model and input configurations shared across the main experiments.}
\label{tab:model_config}
\small
\begin{tabular}{p{5.2cm}p{7.2cm}}
\toprule
\textbf{Item} & \textbf{Value} \\
\midrule
Vision backbone & DINOv2-S~\cite{Oquab2023DINOv2} \\
Backbone type & Vision Transformer \\
Backbone adaptation & LoRA adapters on the pre-trained backbone \\
LoRA rank & 32 \\
Input camera views & 4 \\
Input image resolution & 588 $\times$ 980 per view \\
Patch size & 14 \\
Patch-grid size & 42 $\times$ 70 per view \\
Scene tokens (per view / total) & 16 / 64 \\
Scene-token feature dimension & 256 \\
FFN hidden dimension & 1024 \\
Model parameters & 68.1M \\
Scoring head & Shared six-component scoring head \\
Predicted horizon / poses $T$ & 5s / 10 \\
Inference pipeline & Same as the underlying baseline planner \\
\bottomrule
\end{tabular}
\end{table}

\subsection{Ensemble Details}
\label{app:ensemble}
The NTR (Ensemble) entry in Table~\ref{tab:waymo_short} combines two identical NTR models that differ only in their random seeds (seeds 2 and 3), whose single-model test RFS are 7.9982 and 7.9445.
Following RAP~\cite{feng2025rap}, we pool the candidate trajectories of both models and select the final trajectory with the predicted scores.
The ensemble roughly doubles the inference cost.
All other results in the paper, including the ablations, use a single model.

\subsection{Planning Loss}
\label{app:planning loss}
The planning objective has three terms with all weights
following the base planner:
\begin{equation}
\mathcal{L}_{\mathrm{plan}}
=
\lambda_{\mathrm{traj}}\mathcal{L}_{\mathrm{traj}}
+
\lambda_{\mathrm{score}}\mathcal{L}_{\mathrm{score}}
+
\lambda_{\mathrm{div}}\mathcal{L}_{\mathrm{div}}.
\end{equation}

\paragraph{Trajectory regression} supervises only the candidate closest to the
ground-truth trajectory $\tau^{*}$ (winner-takes-all),
\begin{equation}
\mathcal{L}_{\mathrm{traj}}
=
\min_{i}
\frac{1}{T}
\sum_{t=1}^{T}
\lVert \hat{\tau}_{i,t}-\tau^{*}_{t}\rVert_{1}.
\end{equation}

\paragraph{Diversity regularization} spreads the proposals by penalizing the closest
pair, $\mathcal{L}_{\mathrm{div}}=-\min_{i\neq j}\mathrm{Dist}(\hat{\tau}_i,\hat{\tau}_j)$,
where $\mathrm{Dist}$ is the average $\ell_1$ distance between two trajectories.

\paragraph{Score supervision} trains a shared six-component head
(NC, DAC, DDC, TTC, EP, C) with binary cross-entropy on the benchmark sub-scores,
\begin{equation}
\mathcal{L}_{\mathrm{score}}
=
\sum_{c\in\mathcal{C}}
\mathrm{BCEWithLogits}\!\left(\hat{s}_c, s_c^{*}\right).
\end{equation}
On Waymo we train the same head against RFS-aligned targets rather than a single
scalar RFS, which keeps one scoring interface across both benchmarks.

\section{Architecture Details}
\label{app:architecture}

\subsection{Planner Decoder Details}
\label{app:planner}
This section details the downstream planner decoder $\mathcal{D}_{\phi}$ introduced in Section~\ref{sec:preliminaries}. 
As in the main text, we describe the decoder for a single planning frame and omit the current time index. 
The decoder receives the compact scene-token memory $\tilde{\mathbf{S}}\in\mathbb{R}^{VN_s\times D}$ and ego state $e=[v_{\mathrm{ego}},a,\psi,c]$, and predicts $N_q$ candidate future trajectories over a horizon of $T$ steps:
$\mathcal{D}_{\phi}(\tilde{\mathbf{S}},e)=\{(\hat{\tau}_k,\hat{s}_k)\}_{k=1}^{N_q}$, where $\hat{\tau}_k\in\mathbb{R}^{T\times 3}$ and $\hat{s}_k\in\mathbb{R}$. 
Each candidate trajectory is represented as $\hat{\tau}_k=\{\hat{\mathcal{Y}}_{\tau}^{k}\}_{\tau=1}^{T}$, with each waypoint $\hat{\mathcal{Y}}_{\tau}^{k}=(\hat{x}_{\tau}^{k},\hat{y}_{\tau}^{k},\hat{\psi}_{\tau}^{k})$. 
The corresponding ground-truth trajectory is denoted by $\mathbf{Y}=\{\mathcal{Y}_{\tau}\}_{\tau=1}^{T}$, where $\mathcal{Y}_{\tau}=(x_{\tau},y_{\tau},\psi_{\tau})$.

The ego state is first encoded into a condition embedding:
\begin{equation}
\mathbf{z}_{e}
=
\mathrm{MLP}_{e}(e)
\in\mathbb{R}^{D}.
\end{equation}
The trajectory generation branch maintains $N_q$ learnable trajectory queries:
\begin{equation}
\mathbf{Q}^{\mathrm{traj}}
=
[
\mathbf{q}^{\mathrm{traj}}_1;
\dots;
\mathbf{q}^{\mathrm{traj}}_{N_q}
]
\in
\mathbb{R}^{N_q\times D}.
\end{equation}
After adding the ego-state condition, these queries cross-attend to the compact scene token memory:
\begin{equation}
\mathbf{H}^{\mathrm{traj}}
=
\mathrm{CrossAttn}
\left(
\mathbf{Q}^{\mathrm{traj}}+\mathbf{z}_{e},
\tilde{\mathbf{S}},
\tilde{\mathbf{S}}
\right),
\qquad
\mathbf{H}^{\mathrm{traj}}\in\mathbb{R}^{N_q\times D},
\end{equation}
where $\mathbf{z}_{e}$ is broadcast to all trajectory queries. 
A lightweight trajectory head then decodes the query features into $N_q$ candidate future trajectories:
\begin{equation}
\hat{\boldsymbol{\tau}}
=
h_{\mathrm{traj}}
\left(
\mathbf{H}^{\mathrm{traj}}
\right)
\in
\mathbb{R}^{N_q\times T\times 3}.
\end{equation}

The scoring branch is trajectory-conditioned. Each final proposal is first mapped to a
trajectory embedding by a lightweight MLP over its flattened waypoints, with a
stop-gradient on the trajectory so that the scoring loss does not update the generated trajectory:
\begin{equation}
\mathbf{e}^{\mathrm{traj}}_k=\mathrm{MLP}_{\mathrm{traj}}\!\left(\mathrm{sg}(\hat{\boldsymbol{\tau}}_k)\right)\in\mathbb{R}^{D},
\qquad k=1,\dots,N_q,
\end{equation}
where $\mathrm{sg}(\cdot)$ denotes stop-gradient. The stacked trajectory embeddings
$\mathbf{E}^{\mathrm{traj}}=[\mathbf{e}^{\mathrm{traj}}_1;\dots;\mathbf{e}^{\mathrm{traj}}_{N_q}]$
serve directly as the cross-attention queries onto the same scene-token memory, and the ego
condition is injected after attention:
\begin{align}
\mathbf{H}_{\mathrm{score}}&=\mathrm{CrossAttn}\!\left(\mathbf{E}^{\mathrm{traj}},\,\tilde{\mathbf{S}},\,\tilde{\mathbf{S}}\right),\\
\mathbf{H}_{\mathrm{score}}&\leftarrow\mathbf{H}_{\mathrm{score}}+\mathbf{z}_e.
\end{align}
A lightweight MLP then predicts six sub-scores per candidate:
\begin{equation}
\hat{\mathbf{P}}=\mathrm{MLP}_{\mathrm{score}}(\mathbf{H}_{\mathrm{score}})\in\mathbb{R}^{N_q\times 6},
\end{equation}
corresponding to no-at-fault collision (NC), drivable-area compliance (DAC), driving-direction
compliance (DDC), time-to-collision (TTC), ego progress (EP), and comfort.

Candidate trajectories are ranked using a PDMS-style composition of the predicted sub-scores.
The NC, DAC, and DDC terms are treated as multiplicative penalties, while TTC, EP, and comfort are combined linearly with benchmark-specific weights.
For numerical stability, we compute the composed ranking score in log space:
\begin{equation}
\hat{s}_{k}
=
\sum_{c\in\mathcal{C}_{\mathrm{mul}}}
w_c \log \hat{p}_{k,c}
+
\log
\left(
\sum_{c\in\mathcal{C}_{\mathrm{lin}}}
w_c \hat{p}_{k,c}
\right),
\end{equation}
where
\begin{equation}
\mathcal{C}_{\mathrm{mul}}
=
\{\mathrm{NC},\mathrm{DAC},\mathrm{DDC}\},
\qquad
\mathcal{C}_{\mathrm{lin}}
=
\{\mathrm{TTC},\mathrm{EP},\mathrm{Comfort}\}.
\end{equation}

The final trajectory is selected according to the composed ranking score:
\begin{equation}
k^{*}
=
\arg\max_{k}
\hat{s}_{k},
\qquad
\hat{\tau}
=
\hat{\tau}_{k^{*}}.
\end{equation}

Following the inference-time weighting strategy of the base planner~\cite{kirby2026drivor}, we use benchmark-specific weights to compose the predicted sub-scores.
Table~\ref{tab:inference_score_weights} summarizes the weights used for NavSim v1 and NavSim v2.
For NavSim v1, DDC is disabled by setting its weight to zero, consistent with the PDMS formulation.
For NavSim v2, we use the reweighted configuration adopted by the base planner for EPDMS evaluation.

\begin{table}[htbp]
\centering
\caption{Inference-time weights used to compose predicted trajectory sub-scores.}
\label{tab:inference_score_weights}
\begin{tabular}{lcccccc}
\toprule
Benchmark & NC & DAC & DDC & TTC & EP & Comfort \\
\midrule
Waymo \& NavSim v1 & 1 & 1 & 0 & 5 & 5 & 2 \\
NavSim v2 & 10 & 13 & 6 & 14 & 15 & 2 \\
\bottomrule
\end{tabular}
\end{table}

\subsection{NTR Reconstruction Decoder Details}
\label{app:ntr_decoder}

The reconstruction decoder $g_{\omega}$ is a lightweight Transformer decoder used only during training.
It predicts teacher features at the selected reconstruction locations $\Omega_{\mathrm{rec}}^{v}$.
The input queries are initialized by a shared learnable reconstruction query and the positional encodings of the selected patches:
\begin{equation}
\mathbf{Q}_{\Omega_{\mathrm{rec}}^{v}}^{0}
=
\mathbf{1}_{|\Omega_{\mathrm{rec}}^{v}|}
\mathbf{q}_{\mathrm{rec}}^{\top}
+
\mathbf{E}_{\mathrm{pos}}[\Omega_{\mathrm{rec}}^{v}].
\end{equation}

For decoder layer $\ell=1,\dots,L_{\mathrm{dec}}$, the query features are updated by query self-attention, cross-attention, and a feed-forward block:
\begin{equation}
\bar{\mathbf{Q}}^{\ell}
=
\mathbf{Q}^{\ell-1}
+
\mathrm{MSA}
\left(
\mathrm{LN}(\mathbf{Q}^{\ell-1})
\right),
\end{equation}
\begin{equation}
\hat{\mathbf{Q}}^{\ell}
=
\bar{\mathbf{Q}}^{\ell}
+
\mathrm{MCA}
\left(
\mathrm{LN}(\bar{\mathbf{Q}}^{\ell}),
\mathrm{LN}(\tilde{\mathbf{S}}^{v,m})
\right),
\end{equation}
\begin{equation}
\mathbf{Q}^{\ell}
=
\hat{\mathbf{Q}}^{\ell}
+
\mathrm{FFN}
\left(
\mathrm{LN}(\hat{\mathbf{Q}}^{\ell})
\right).
\end{equation}
Here, $\mathrm{MSA}$ denotes query self-attention and $\mathrm{MCA}$ denotes multi-head cross-attention. 
The keys and values in cross-attention are computed only from the masked-view scene tokens $\tilde{\mathbf{S}}^{v,m}$, and the dense online patch outputs are never used as decoder memory.

The final decoder state is projected to the teacher feature dimension by a lightweight prediction head $h_{\omega}$:
\begin{equation}
\hat{\mathbf{Z}}_{\Omega_{\mathrm{rec}}^{v}}^{v}
=
h_{\omega}
\left(
\mathbf{Q}^{L_{\mathrm{dec}}}
\right).
\end{equation}
The reconstruction loss for view $v$ is
\begin{equation}
\mathcal{L}_{\mathrm{rec}}^{v}
=
\frac{1}{|\Omega_{\mathrm{rec}}^{v}|}
\sum_{i\in\Omega_{\mathrm{rec}}^{v}}
\left\|
\hat{\mathbf{Z}}_{i}^{v}
-
\mathbf{Z}_{i}^{v,*}
\right\|_{1},
\end{equation}
and the final reconstruction loss is averaged over all camera views. 
Because the decoder cross-attends only to scene tokens, the reconstruction objective directly supervises the scene-token bottleneck rather than the dense online patch outputs.

\paragraph{Reconstruction decoder configuration.}
$g_{\omega}$ is an $L_{\mathrm{dec}}$-layer pre-norm Transformer decoder, where each layer applies query self-attention ($\mathrm{MSA}$), cross-attention ($\mathrm{MCA}$) onto the masked-view scene tokens $\tilde{\mathbf{S}}^{v,m}$, and a feed-forward block. 
The default configuration is summarized in Table~\ref{tab:recon_decoder}. 
The decoder is deliberately lightweight relative to the planner, since its only role is to route the dense reconstruction gradient through the scene-token bottleneck, and it is discarded after training.

\begin{table}[htbp]
\centering
\caption{Reconstruction decoder $g_{\omega}$ configuration (training only).}
\label{tab:recon_decoder}
\begin{tabular}{ll}
\toprule
Item & Value \\
\midrule
Decoder layers $L_{\mathrm{dec}}$       & 4 \\        
Hidden dimension                        & 256 \\      
Attention heads                         & 8 \\        
FFN hidden dimension                    & 1024 \\
Cross-attention memory                  & masked-view scene tokens $\tilde{\mathbf{S}}^{v,m}$ only \\
Query init.                             & shared learnable query $+$ patch pos.\ enc. \\
Prediction head $h_{\omega}$            & single linear layer to teacher dim. \\
\bottomrule
\end{tabular}
\end{table}

\section{Semantic Prior Generation and Ablations}
\label{app:semantic_prior}
 
\subsection{Prompt Groups, Category Weights, and Pooling}
\label{app:semantic_pipeline}
We generate the semantic priors with SAM3~\cite{carion2026sam} as a promptable
segmentation model. For each camera view we query SAM3 with four prompt groups
(vehicles, humans, traffic lights, and road regions). When a group returns
several instance masks we merge them by binary union into one category mask
$U^{(k)}_v=\bigcup_n U^{(k,n)}_v$. Stacking the category masks gives
$U\in\{0,1\}^{V\times K\times H\times W}$, which we average-pool to the planner
patch grid $G=\mathrm{Pool}(U)\in\mathbb{R}^{V\times K\times H_p\times W_p}$ and
normalize along the semantic dimension, so $G(v,k,i,j)$ is the prior score of
category $k$ at patch $(i,j)$. These priors only allocate reconstruction targets
during training. They add no perception output and are removed together with the
reconstruction branch at inference. Algorithm~\ref{alg:semantic_prior} gives the
full procedure.
 
\begin{algorithm}[t]
\caption{Prompt-Based Semantic Prior Generation and Reconstruction Target Allocation}
\label{alg:semantic_prior}
\begin{algorithmic}[1]
\Require Multi-view images $\mathcal{I}=\{I^v\}_{v=1}^{V}$; prompt set $\mathcal{P}=\{p_k\}_{k=1}^{K}$; foundation segmentation model $\Phi$; patch-grid size $(H_p,W_p)$; reconstruction ratio $\rho_{\mathrm{rec}}$; category weights $\{\alpha_k\}_{k=1}^{K}$; noise temperature $\tau$
\Ensure Reconstruction target sets $\{\Omega_{\mathrm{rec}}^v\}_{v=1}^{V}$
 
\For{$v=1,\dots,V$}
    \For{$k=1,\dots,K$}
        \State Query $\Phi$ with image $I^v$ and prompt $p_k$
        \State Obtain image-level mask $\mathbf{U}(v,k,:,:)$
    \EndFor
\EndFor
 
\State Pool masks to the planner patch grid:
\[
\mathbf{G}
\gets
\mathrm{Pool}(\mathbf{U})
\in
\mathbb{R}^{V\times K\times H_p\times W_p}
\]
 
\State Normalize $\mathbf{G}$ along the semantic dimension
 
\State Initialize full-grid random scores:
\[
\mathbf{A}(v,i,j)
\sim
\mathcal{U}(0,1)
\]
 
\If{semantic priors are available}
    \State Compute semantic-biased reconstruction scores:
    \[
    \mathbf{A}(v,i,j)
    \gets
    \sum_{k=1}^{K}
    \alpha_k\,\mathbf{G}(v,k,i,j)
    +
    \tau\,\varepsilon_{vij},
    \qquad
    \varepsilon_{vij}\sim\mathcal{N}(0,1)
    \]
\EndIf
 
\State Set the reconstruction budget:
\[
m
\gets
\left\lfloor
\rho_{\mathrm{rec}}H_pW_p
\right\rfloor
\]
 
\For{$v=1,\dots,V$}
    \State Select the top-scoring patches over the full patch grid:
    \[
    \Omega_{\mathrm{rec}}^v
    \gets
    \mathrm{Top}\text{-}m
    \left(
    \mathbf{A}(v,:,:)
    \right)
    \]
\EndFor
 
\State \Return $\{\Omega_{\mathrm{rec}}^v\}_{v=1}^{V}$
\end{algorithmic}
\end{algorithm}
 
The priors are a soft bias over the full patch grid, not a hard foreground
constraint. When a view has no valid semantic region the semantic term is zero,
selection falls back to the uniform random scores, and the procedure recovers the
random-masking baseline used in our ablations.

\paragraph{Category weights from area statistics.}
The weights $\alpha$ control how the reconstruction budget is split across
categories, not how important each category is for planning, and we set them from
a simple dataset statistic rather than by tuning. On the training set the
structured foreground covers about $31\%$ of an image, and road and vehicles make
up almost all of it ($70.8\%$ and $28.2\%$), while pedestrians and traffic lights
together account for about $1\%$ (Table~\ref{tab:sem_stats}). With equal weights the selected targets would be
almost entirely road, and small but safety-critical structures would rarely be
picked. We therefore halve the road weight and leave the other three categories
at $1.0$,
\[
(\alpha_{\text{veh}},\alpha_{\text{hum}},\alpha_{\text{ctrl}},\alpha_{\text{road}})
=(1.0,\,1.0,\,1.0,\,0.5).
\]
Down-weighting road only offsets its size and does not mean road geometry matters
less for planning. We keep $\alpha$ fixed across all benchmarks; these statistics
justify the direction of the reweighting rather than an optimal setting, and we
study its sensitivity in Section~\ref{app:prior_sensitivity}.
 
\begin{table}[htbp]
\centering
\caption{
Category distribution within the structured foreground on the Waymo E2E training set. We applied the fixed allocation weight $\alpha$ assigned to each category.
}
\label{tab:sem_stats}
\begin{tabular}{lcc}
\toprule
Category & Share of structured foreground & Weight $\alpha$ \\
\midrule
Road            & 70.8\% & 0.5 \\
Vehicle         & 28.2\% & 1.0 \\
Pedestrian      &  0.6\% & 1.0 \\
Traffic light   &  0.4\% & 1.0 \\
\bottomrule
\end{tabular}
\end{table}


\subsection{Masking Strategy}
\label{app:mask_strategy}

\begin{figure}[htbp]
\centering
\includegraphics[width=\linewidth]{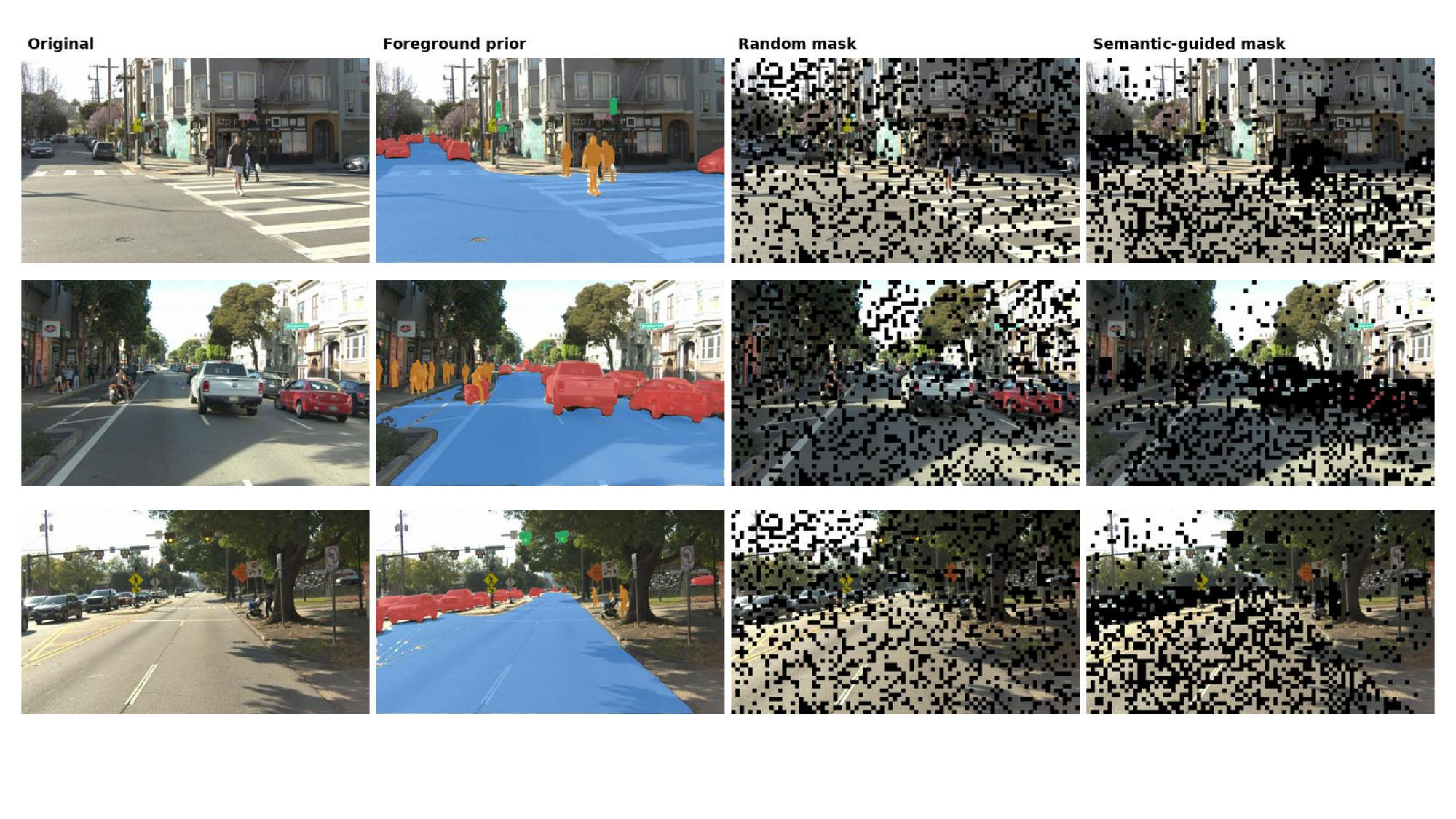}
\caption{Reconstruction target allocation under different masking strategies,
with selected patches shown as overlays. Compared with uniform masking, the
semantic prior concentrates the reconstruction budget on agents and
traffic-control cues instead of sky or irrelevant background regions.}
\label{fig:mask_strategy}
\end{figure}

Table~\ref{tab:core_ablation} of the main paper compares random and semantic-prior selection
quantitatively. Here we further visualize where each strategy places its
reconstruction targets. Figure~\ref{fig:mask_strategy} overlays the selected
patches on three scenes under uniform masking and semantic-prior masking under
different reconstruction mask ratios $\rho_{\mathrm{rec}}$. Uniform masking spends
much of the budget on sky, distant background, and road texture, while the
semantic prior places more targets on vehicles, pedestrians, and traffic lights,
while still preserving enough road coverage for geometry.

\paragraph{Reconstruction ratio.}
The reconstruction ratio $\rho_{\mathrm{rec}}$ sets how many patches are masked and reconstructed. A small
ratio leaves the bottleneck barely constrained, while a large ratio makes
reconstruction too hard for a handful of scene tokens to solve and lets a few
difficult patches dominate the gradient. Table~\ref{tab:rho} reports how planning
responds to $\rho_{\mathrm{rec}}$ with fixed $\tau = 0.3$, which is best at an intermediate value.
\begin{table}[htbp]
\centering
\caption{Sensitivity of planning to the reconstruction ratio $\rho_{\mathrm{rec}}$
(Waymo~E2E validation split).}
\label{tab:rho}
\begin{tabular}{lccc}
\toprule
$\rho_{\mathrm{rec}}$ & RFS$\uparrow$ & ADE@5s$\downarrow$ & ADE@3s$\downarrow$ \\
\midrule
0.2 & 7.936 & 2.217 & 1.018 \\ 
0.3 & \textbf{7.974} & \textbf{2.146} & \textbf{0.949} \\ 
0.4 & 7.943 & 2.281 & 1.087 \\ 
0.5 & 7.890 & 2.309 & 1.098\\
\bottomrule
\end{tabular}
\end{table}
 
 

\subsection{Sensitivity to the Semantic Prior}
\label{app:prior_sensitivity}
Semantic priors provide only a soft allocation bias under a fixed reconstruction
budget rather than defining all planning-relevant regions, and the category weights
$[\alpha_{\mathrm{veh}},\alpha_{\mathrm{hum}},\alpha_{\mathrm{ctrl}},\alpha_{\mathrm{road}}]$
only redistribute this fixed budget across semantic groups.
Table~\ref{tab:prior_sensitivity} varies the prior generator, the prompt coverage
and the category weights, keeping all other settings at their defaults.
Replacing SAM3 with Grounded-SAM~\cite{ren2024grounded} gives nearly identical
performance (7.969 vs.\ 7.974 RFS), so the gain does not depend on a specific
foundation model. All prompt and weight variants improve over random allocation;
removing actor prompts causes the largest drop among prompt variants, and the
control-heavy weighting, which over-allocates the budget to small traffic-control
regions, performs worst among weight variants. Removing the road prompt slightly
improves RFS (8.006); we keep the full prompt set as the default since it was fixed
before this study rather than tuned on the validation split.

\begin{table}[htbp]
\centering
\caption{Sensitivity to the semantic prior on the Waymo E2E validation split.
Weights are listed as $[\alpha_{\mathrm{veh}},\alpha_{\mathrm{hum}},\alpha_{\mathrm{ctrl}},\alpha_{\mathrm{road}}]$.}
\label{tab:prior_sensitivity}
\begin{tabular}{llc}
\toprule
Axis & Setting & RFS $\uparrow$ \\
\midrule
No prior & Random allocation & 7.817 \\
\midrule
Generator & Grounded-SAM & 7.969 \\
\midrule
\multirow{3}{*}{Prompt}
& w/o actors $[0,0,1,0.5]$ & 7.862 \\
& w/o controls $[1,1,0,0.5]$ & 7.922 \\
& w/o road $[1,1,1,0]$ & \textbf{8.006} \\
\midrule
\multirow{3}{*}{Weights}
& Equal $[1,1,1,1]$ & 7.934 \\
& Actor-heavy $[10,10,1,0.5]$ & 7.951 \\
& Control-heavy $[1,1,10,0.5]$ & 7.826 \\
\midrule
Default & SAM3 + full $[1,1,1,0.5]$ & 7.974 \\
\bottomrule
\end{tabular}
\end{table}

\section{Scene-Token Diagnostics}
\label{app:additional_analysis}

The main paper already quantifies scene-token redundancy across token budgets
(Table~5). Here we provide complementary qualitative analyses of how the compact
memory is used, and revisit the redundancy metrics (cosine similarity, effective
rank) in a controlled comparison against a direct decorrelation penalty
(Section~\ref{app:diag_penalty}).

\paragraph{Diagnostic metrics.}
We measure scene-token redundancy within each camera view.
For a given sample and camera view $v$, let
$\{t_{v,n}\}_{n=1}^{N_s}$ denote the $N_s$ scene-token features associated with that view.
We compute the metrics independently for each camera view and then average them over views and evaluation samples.

The pairwise cosine similarity is defined as
\[
\mathrm{Sim}^{v}
=
\frac{2}{N_s(N_s-1)}
\sum_{a<b}
\frac{
t_{v,a}^{\top}t_{v,b}
}{
\lVert t_{v,a}\rVert
\lVert t_{v,b}\rVert
}.
\]
It measures the redundancy among scene tokens extracted from the same camera view, with lower values indicating less overlapping representations.

We also compute the effective rank of the per-view token matrix
$T^{v}=[t_{v,1};\dots;t_{v,N_s}]\in\mathbb{R}^{N_s\times D}$:
\[
\mathrm{EffRank}(T^{v})
=
\exp\!\Big(
-\sum_i p_i\log p_i
\Big),
\qquad
p_i
=
\frac{\sigma_i}{\sum_j\sigma_j},
\]
where $\{\sigma_i\}$ are the singular values of $T^{v}$.
A higher effective rank indicates that the compact scene-token memory spans more independent feature directions.




\subsection{Scene-Token Similarity Across Camera Views}
\label{app:diag_similarity}

\begin{figure*}[htbp]
\centering
\includegraphics[width=\textwidth]{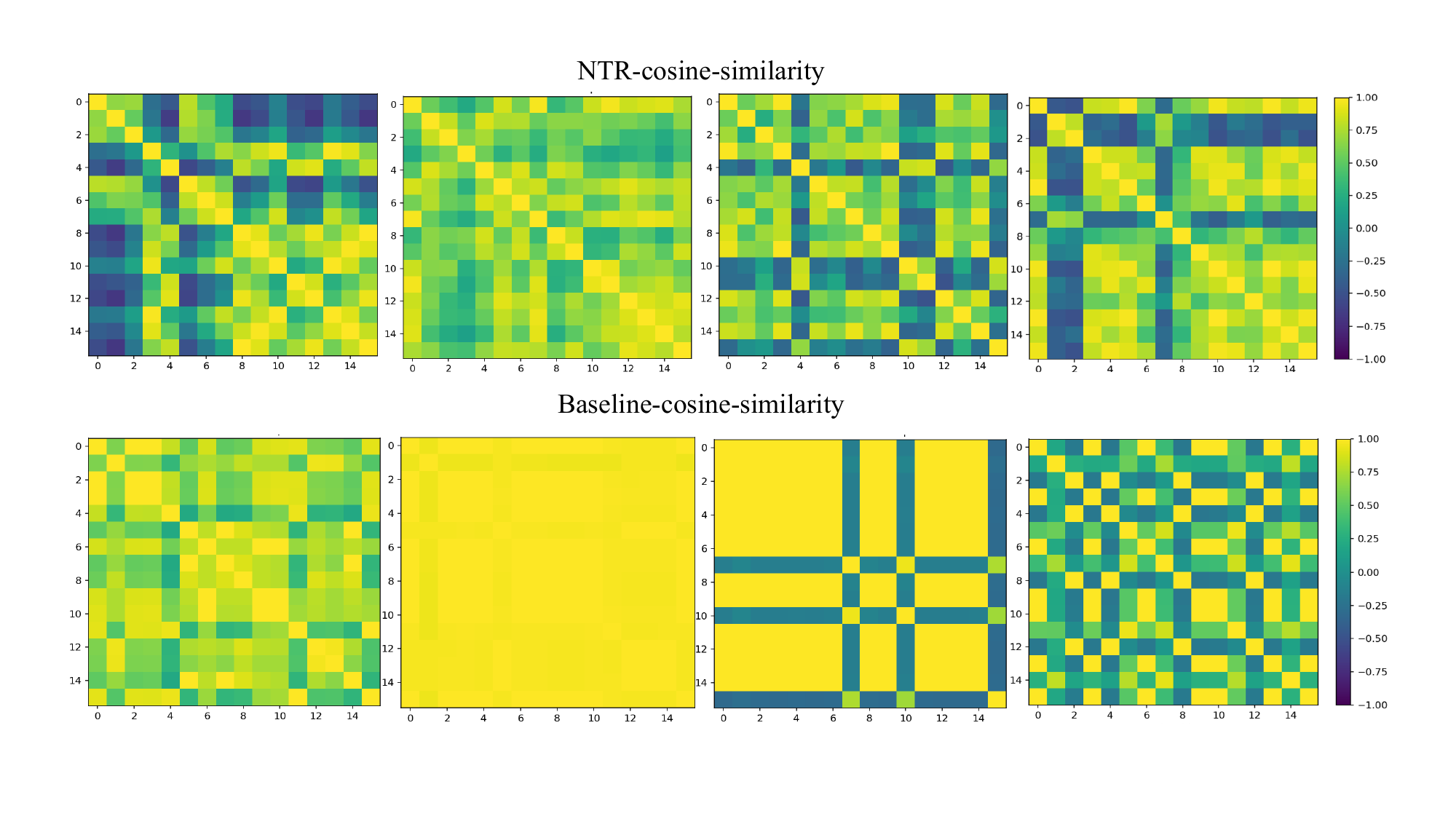}
\caption{
Pairwise cosine similarity of scene tokens across four camera views (front-right, front, front-left, and rear).
For each view, we compare the planning-only baseline and NTR under the standard 16-token setting.
Darker values indicate lower cosine similarity.
NTR consistently produces less redundant scene-token representations across camera views.
}
\label{fig:token_similarity_views}
\end{figure*}

We further compare the pairwise cosine similarity of scene tokens across all four camera views.
Figure~\ref{fig:token_similarity_views} visualizes the $16\times16$ token-similarity matrices of the planning-only baseline and NTR for the front-right, front, front-left, and rear cameras on Waymo E2E validatioin set.
The planning-only baseline exhibits consistently high inter-token similarity, indicating that multiple scene tokens encode redundant visual information.
In contrast, NTR substantially reduces token similarity across camera views, suggesting that direct bottleneck supervision encourages a more differentiated use of the compact scene-token memory.

\subsection{Per-Token Attention Maps}
\label{app:diag_attn}

\begin{figure*}[p]
\centering

\includegraphics[width=\linewidth,trim={0.8cm 4.5cm 0.8cm 3.8cm},clip]{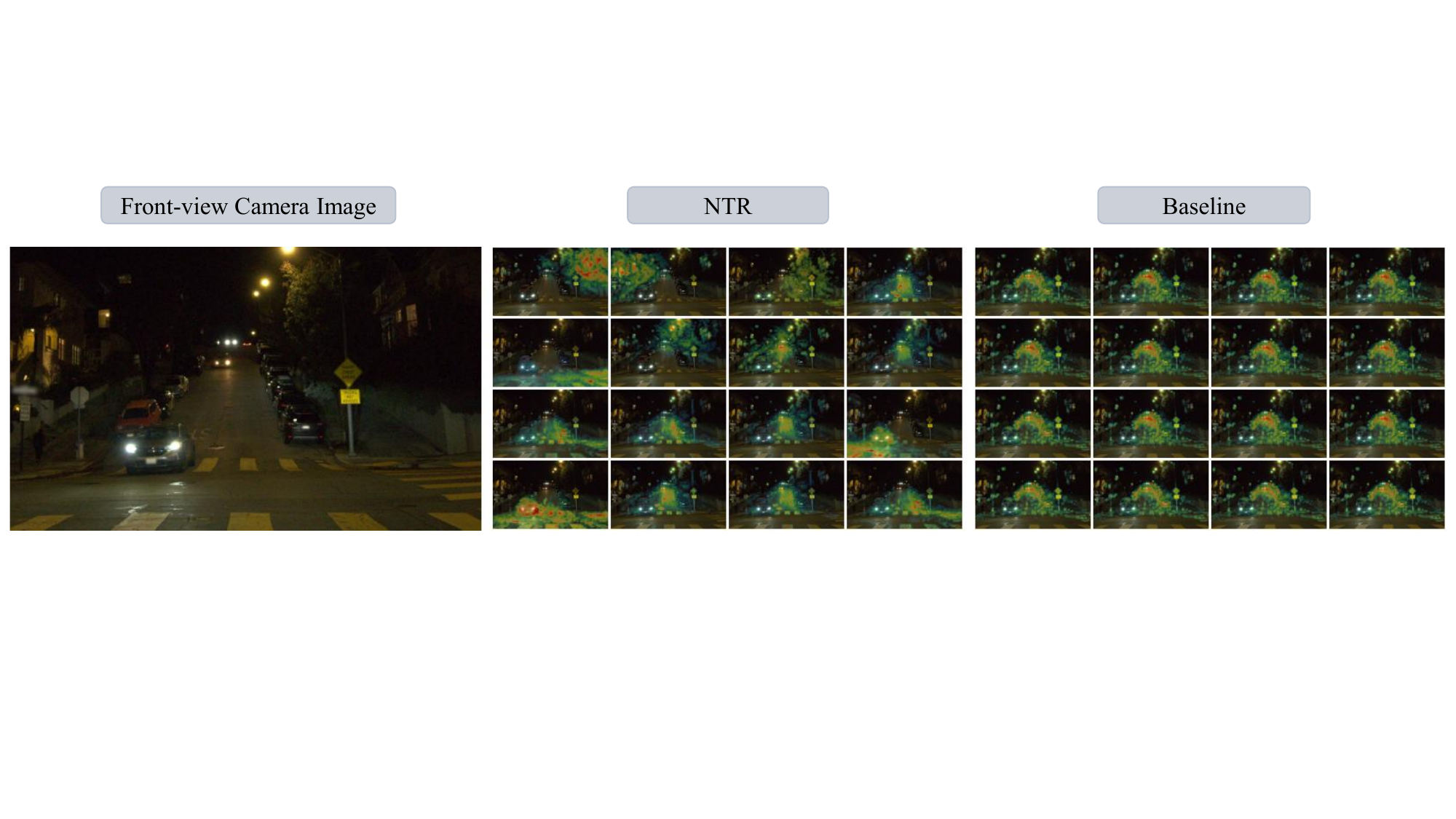}

\vspace{0.3em}
\includegraphics[width=\linewidth,trim={0.8cm 4.5cm 0.8cm 3.8cm},clip]{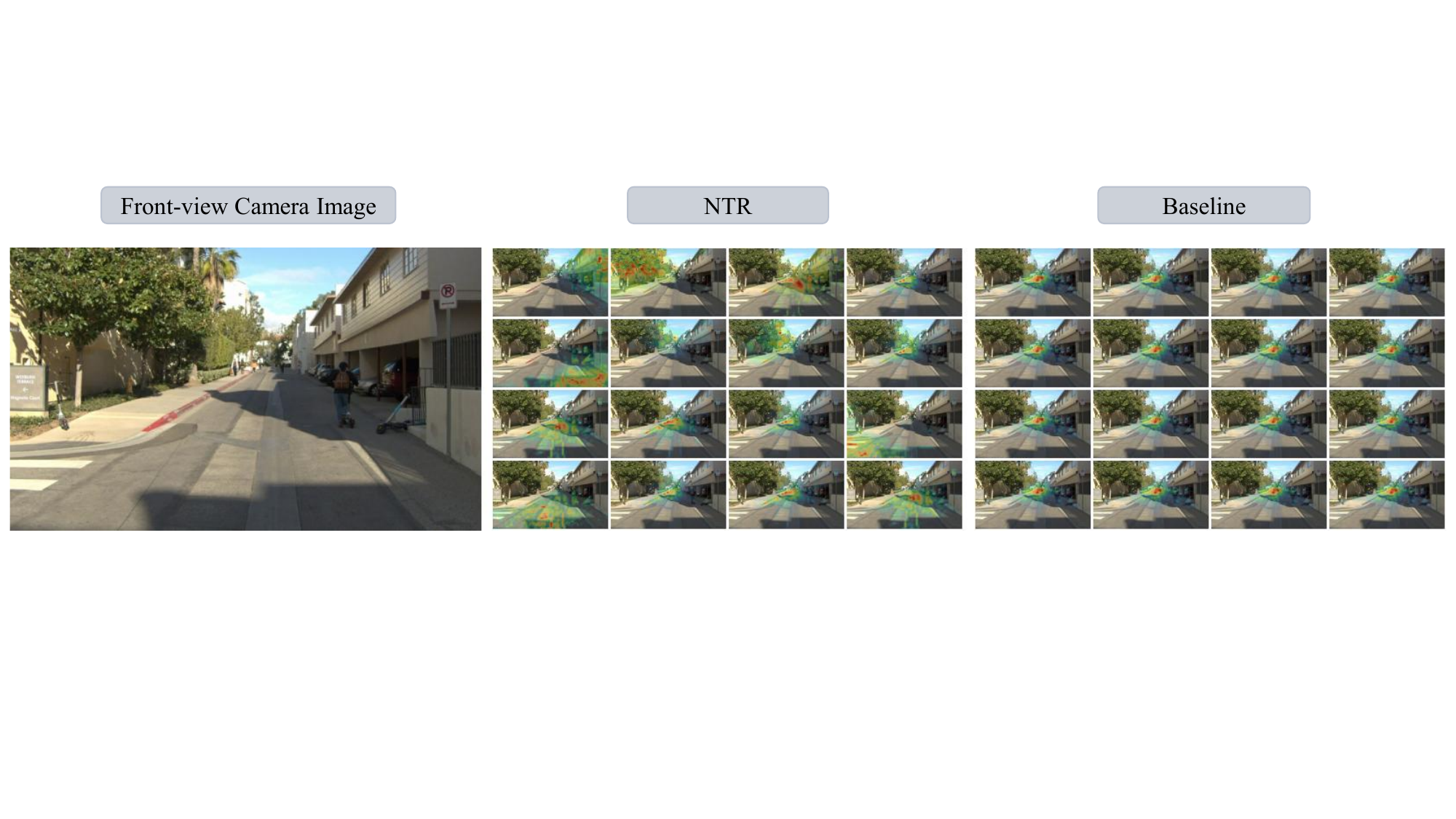}

\vspace{0.3em}
\includegraphics[width=\linewidth,trim={0.8cm 4.5cm 0.8cm 3.8cm},clip]{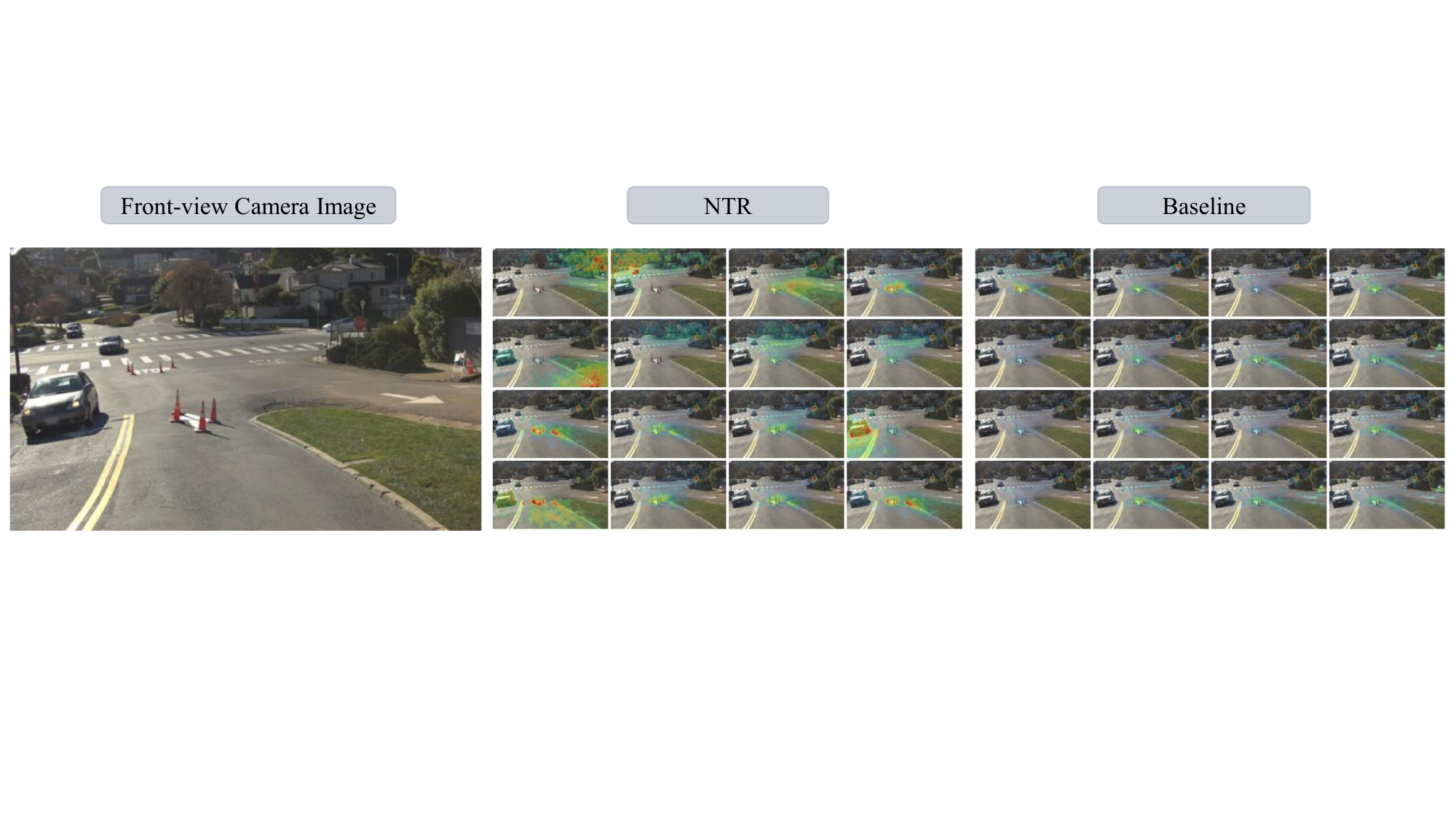}

\caption{
Per-token attention maps across three driving scenes.
For each example, we show the input front-view image together with the 16 scene-token attention maps of the planning-only baseline and NTR.
Compared with the baseline, NTR shows more differentiated token responses and broader coverage over driving-relevant regions, while the baseline exhibits more overlapping attention patterns.
}
\label{fig:scene_token_all}
\end{figure*}

Figure~\ref{fig:scene_token_all} visualizes the attention maps of all 16 scene tokens for the same front-view camera image across three driving scenes.
Each example includes the original front-view image, the $4\times4$ per-token attention maps of NTR, and the corresponding attention maps of the planning-only baseline.

In the baseline, the per-token maps are highly correlated: multiple tokens tend
to attend to similar dominant regions, leaving other parts of the scene
under-covered. Under NTR, the tokens become more differentiated and attend to
more diverse driving-relevant regions, including agents, lane boundaries, and
traffic-control cues. This provides a spatial view of the lower token redundancy
shown by the cosine-similarity analysis in Figure~\ref{fig:token_similarity_views}.




\subsection{PCA of Scene-Token Features}
\label{app:diag_pca}

\begin{figure*}[htbp]
\centering
\includegraphics[width=\textwidth,trim={0.3cm 1.8cm 0.3cm 2.0cm},clip]{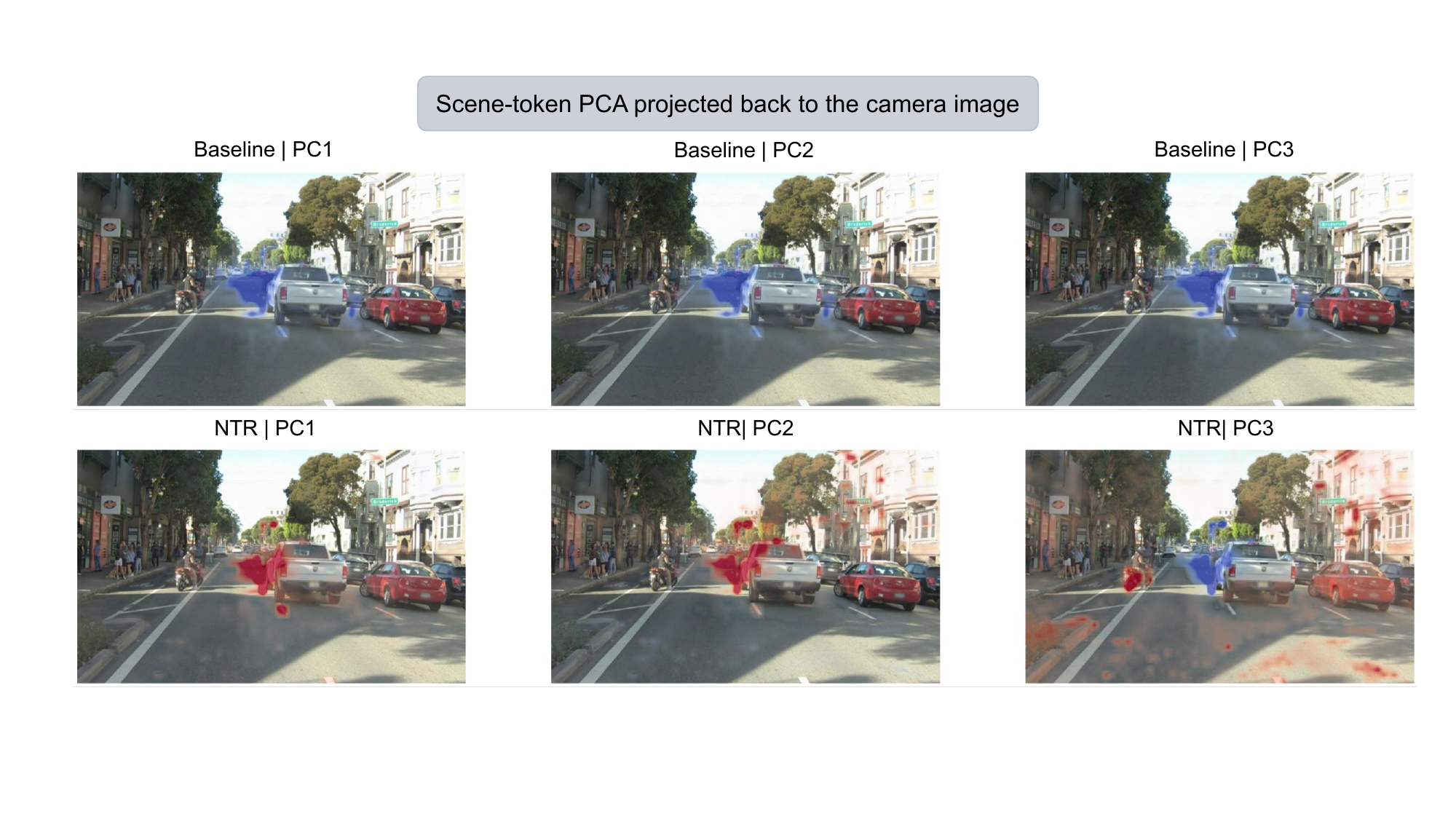}
\caption{
PCA projection of scene-token features back to the front-view camera image.
Compared with the planning-only baseline, NTR produces richer and more spatially
dispersed principal-component responses, consistent with higher effective rank
and lower token redundancy.
}
\label{fig:pca}
\end{figure*}

To further inspect token diversity in feature space, we project the first three
principal components of the $16$ scene-token features back to the front-view
camera image. Figure~\ref{fig:pca} compares the planning-only baseline and NTR
on a matched scene. The baseline principal-component responses are more localized
and visually similar, indicating more redundant token usage. In contrast, NTR
produces more spatially dispersed and differentiated principal-component maps,
suggesting that the scene tokens encode more diverse and complementary visual
evidence. This provides a feature-space view that complements the spatial
attention analysis above.

\subsection{Comparison with Direct Token-Decorrelation Regularization}
\label{app:diag_penalty}

We further examine whether reducing scene-token redundancy alone is sufficient to improve planning.
As a controlled baseline, we add a direct token-decorrelation objective:
\begin{equation}
\mathcal{L}_{\mathrm{pen}}
=
\lambda_{\mathrm{pen}}
\frac{2}{N_s(N_s-1)}
\sum_{a<b}
\left(
\frac{
t_a^\top t_b
}{
\lVert t_a\rVert
\lVert t_b\rVert
}
\right)^2.
\end{equation}
Unlike NTR, this loss directly optimizes token diversity without supervising the visual content retained in the compact memory.

As shown in Table~\ref{tab:penalty}, direct decorrelation substantially reduces token similarity and increases effective rank, but yields only a marginal planning improvement.
NTR achieves stronger planning performance with a more moderate change in the redundancy metrics.
This suggests that reducing token redundancy alone is insufficient. The compact memory must also retain informative visual evidence for planning.

\begin{table}[htbp]
\centering
\caption{
Comparison with direct token-decorrelation regularization on the Waymo E2E validation split.
}
\label{tab:penalty}
\begin{tabular}{lcccc}
\toprule
Method 
& RFS $\uparrow$ 
& ADE@5s $\downarrow$ 
& Eff.\ Rank $\uparrow$ 
& Token Sim. $\downarrow$ \\
\midrule
Planning-only baseline
& 7.652 & 2.565 & 2.080 & 0.953 \\
Baseline $+$ decorrelation penalty
& 7.695 & 2.528 & \textbf{8.165} & \textbf{0.107} \\
NTR
& \textbf{7.974} & \textbf{2.146} & 3.380 & 0.407 \\
\bottomrule
\end{tabular}
\end{table}

\subsection{Attention to Regions Outside the Semantic Prompts}
\label{app:diag_background}
Regions outside the semantic prompts may still contain useful visual context, and
the stochastic term $\tau\varepsilon$ in the allocation score keeps part of the
reconstruction budget on them. We measure how much scene-token attention these
regions receive on the Waymo E2E validation split. They occupy 68.38\% of the image
area and receive 67.42\% of the pooled scene-token attention, and 76.76\% of
token-attention instances assign more than half of their attention to them.
Higher attention to these regions does not correspond to worse planning: the
quartile of samples with the highest such attention has an ADE@5s of 1.825\,m,
compared with 2.301\,m for the lowest quartile. Hence the semantic prior biases
reconstruction without restricting the scene tokens to the prompted categories.

\subsection{Alternative Reconstruction Targets}
\label{app:alt_targets}
We isolate the effect of the reconstruction target representation by replacing the
EMA latent targets while fixing the planner, the semantic allocation and the
scene-token budget. Table~\ref{tab:alt_targets} compares raw RGB patches, latents
of a pretrained VAE (\textit{sd-vae-ft-mse}~\cite{Rombach_2022_CVPR}), frozen DINOv2
features and the EMA teacher used in NTR. Latent targets outperform
pixel targets, and adaptive EMA targets perform best.

\begin{table}[htbp]
\centering
\caption{Effect of the reconstruction target on the Waymo E2E validation split.
All variants use semantic-prior allocation.}
\label{tab:alt_targets}
\begin{tabular}{lc}
\toprule
Reconstruction target & RFS $\uparrow$ \\
\midrule
Planning-only baseline (no reconstruction) & 7.652 \\
RGB patches & 7.739 \\
Frozen DINOv2 features & 7.781 \\
VAE latents (\textit{sd-vae-ft-mse}) & 7.866 \\
EMA teacher features (NTR) & \textbf{7.974} \\
\bottomrule
\end{tabular}
\end{table}

\section{Efficiency Analysis}
\label{app:efficiency}

NTR introduces training-time overhead only.
The EMA teacher and reconstruction decoder are used only during training, while semantic priors are generated offline with SAM3 and cached for reuse.
At inference time, the planner does not invoke any NTR-specific auxiliary components.
The inference graph is therefore identical to the planning-only baseline.
On a single RTX 3090 GPU, the baseline and NTR achieve comparable per-frame latency: 119.2\,ms and 119.8\,ms, respectively.

During training, the baseline contains 41.1M instantiated parameters, including 17.5M trainable parameters.
NTR increases the instantiated parameter count to 68.1M, including 22.1M trainable parameters.
The additional cost primarily comes from the EMA teacher forward pass and the lightweight reconstruction decoder.
On our 8$\times$A800 setup, one epoch takes approximately 20\,min for the baseline and 30--40\,min for NTR.
Generating semantic priors for the full training set is a one-time offline preprocessing step that takes approximately 2 days on 8$\times$A800 GPUs.

\begin{table}[htbp]
\centering
\caption{
Efficiency comparison.
NTR introduces additional training-time cost without changing inference-time computation.
}
\label{tab:efficiency}
\begin{tabular}{lcc}
\toprule
 & Baseline & NTR \\
\midrule
Inference latency (ms/frame)       & 119.2 & 119.8 \\
Training-stage parameters          & 41.1M & 68.1M \\
Trainable parameters               & 17.5M & 22.1M \\
Non-trainable parameters           & 23.6M & 46.0M \\
Training time (min/epoch)          & $\sim$20 & $\sim$30--40 \\
Semantic-prior generation          & -- & 2 days, offline \\
\bottomrule
\end{tabular}
\end{table}

\section{Scaling and Deployment}
\label{app:scaling_deployment}

\subsection{Data Scaling on a Large-Scale Private Dataset}
\label{app:scaling_data}
We evaluate whether the benefits of bottleneck supervision persist as the amount of training data increases, using a large-scale private driving dataset. Since no closed-loop simulator is available for this dataset, this setting is evaluated using open-loop metrics only. Specifically, we report Average Displacement Error (ADE), Final Displacement Error (FDE), and Miss Rate (MR) computed against logged ground truth futures.

Table~\ref{tab:scaling} summarizes the results as the training set is scaled from 100k to 1.6M samples. As expected, both the baseline and NTR benefit from additional training data. However, NTR consistently outperforms the baseline across all data scales, achieving lower displacement errors and lower miss rates. Notably, the gains are particularly pronounced in the low-data regime, suggesting that direct supervision of the scene-token bottleneck improves representation quality and data efficiency. Importantly, the performance gap remains significant even at the largest scale of 1.6M samples, indicating that the benefits of bottleneck supervision are complementary to, rather than replaced by, increasing amounts of training data.
 
\subsection{Real-Vehicle Deployment}
\label{app:scaling_deploy}

We further integrate NTR into a production real-vehicle planning stack to examine its deployability beyond offline benchmarks. Since NTR is used only as an auxiliary representation-learning objective during training, the deployed planner preserves the same inference architecture, input-output interfaces, and runtime execution pipeline as the baseline planner. As a result, integrating NTR does not require modifying the on-vehicle inference engine, scheduling logic, communication protocol, or downstream control interface. This property makes NTR particularly practical for production deployment: the training procedure improves the compact planning representation, while the deployed model remains compatible with the existing vehicle-side inference stack.

During real-vehicle deployment, the planner takes six surround-view camera images as visual input, together with additional navigation route information that provides high-level driving guidance. The NTR-trained planner runs on one edge inference chip and directly outputs planning results for the vehicle stack. In parallel, a second edge inference chip runs a BEV perception model for visualization and monitoring purposes. This BEV branch is used only for visual inspection and does not change the planner inference path, ensuring that the deployed NTR planner remains perception-free and consistent with the benchmark setting.

Deployment videos demonstrating qualitative closed-loop behavior in real-world driving scenarios are available at \projecturl. We present these experiments as deployment validation rather than as a statistical safety evaluation.

\begin{table}[htbp]
\centering
\caption{Open-loop data-scaling on the private dataset.  NTR improves over the baseline at every data scale.}
\label{tab:scaling}
\begin{tabular}{lcccccc}
\toprule
Training & \multicolumn{3}{c}{Baseline} & \multicolumn{3}{c}{NTR} \\
\cmidrule(lr){2-4}\cmidrule(lr){5-7}
samples & ADE$\downarrow$ & FDE$\downarrow$ & MR ($\%$$\downarrow$) & ADE$\downarrow$ & FDE$\downarrow$ & MR ($\%$$\downarrow$)  \\
\midrule
100 k  & 1.462 & 4.138 & 22.46 & 1.379 & 3.924 & 20.53 \\
200 k  & 1.416 & 3.986 & 21.29 & 1.313 & 3.702 & 19.02\\
400 k  & 1.404 & 3.771 & 19.68 & 1.275 & 3.481 & 17.37 \\
800 k  & 1.292 & 3.257 & 16.56 & 1.102 & 3.057 & 15.09\\
1.6 M  & 1.033 & 2.933 & 14.63 & 0.961 & 2.727 & 13.21\\
\bottomrule
\end{tabular}
\end{table}

\section{Additional Qualitative Results and Failure Cases}
\label{app:qualitative}

\subsection{Additional Planning Trajectory Comparison}

\label{app:qual_traj}
\begin{figure*}[!htbp]
\centering
\includegraphics[width=0.7\textwidth,trim={0.2cm 3cm 0.2cm 3cm},clip]{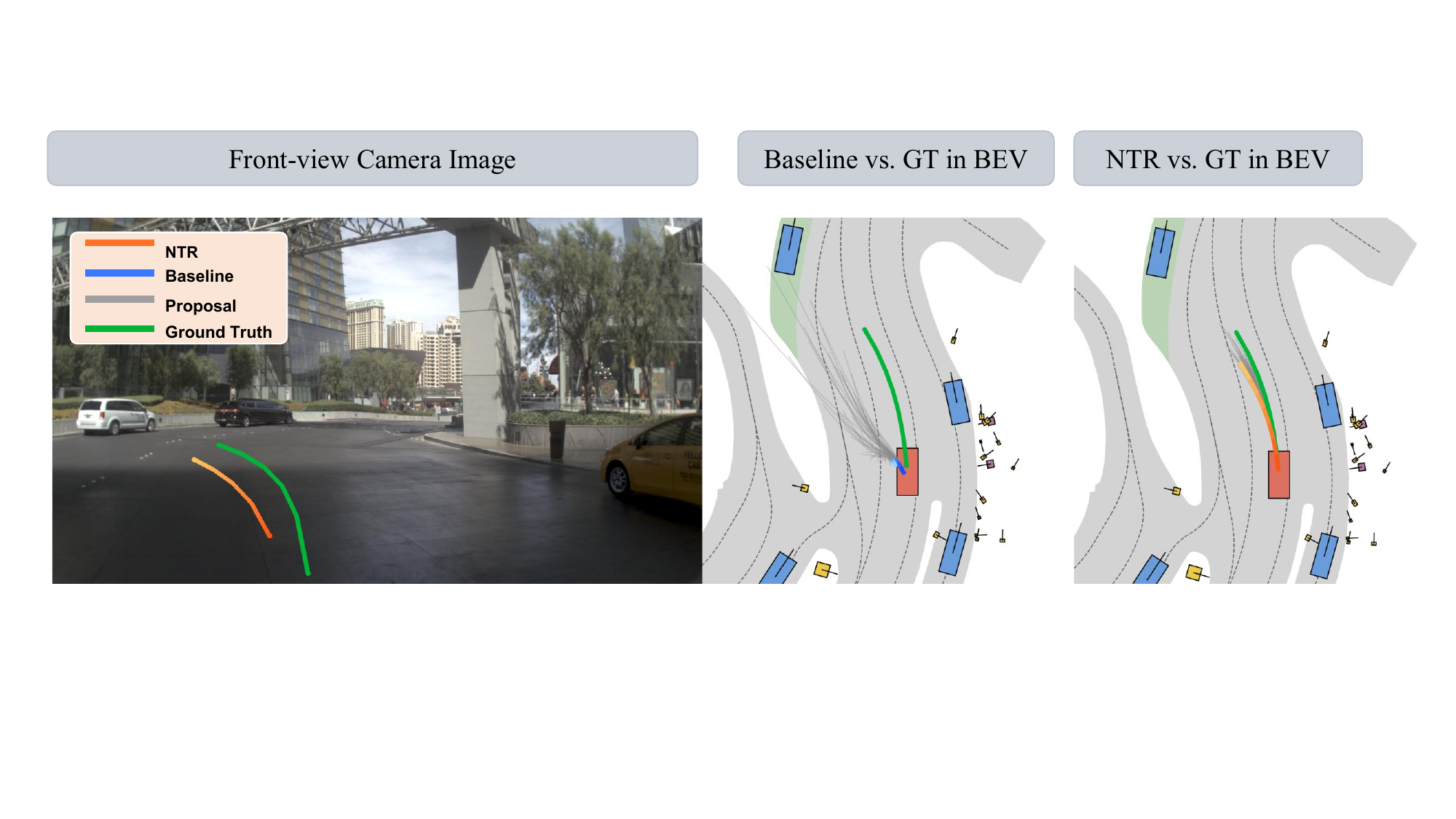}\\[3pt]
\includegraphics[width=0.7\textwidth,trim={0.2cm 3cm 0.2cm 3cm},clip]{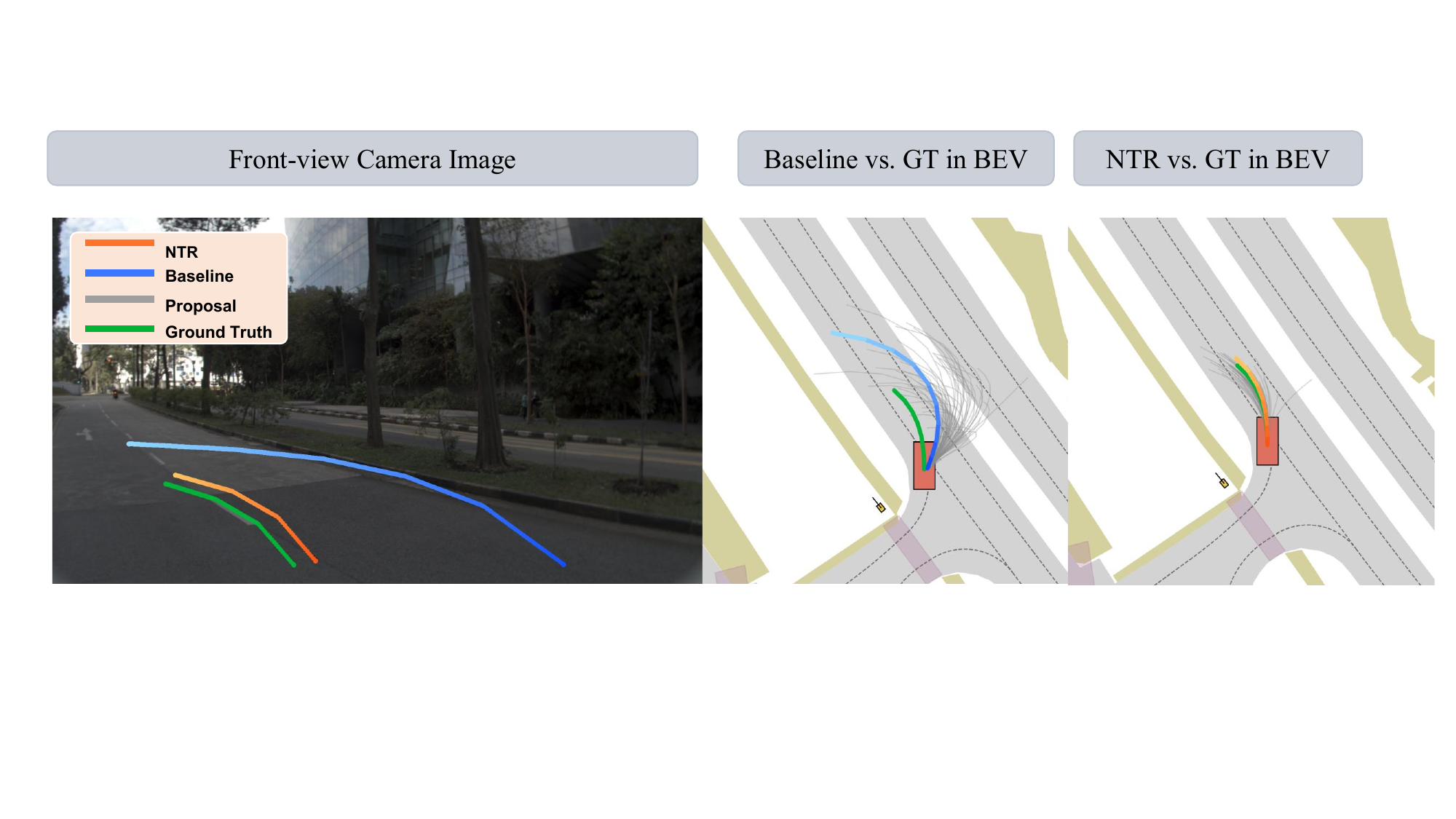}\\[3pt]
\includegraphics[width=0.7\textwidth,trim={0.2cm 3cm 0.2cm 3cm},clip]{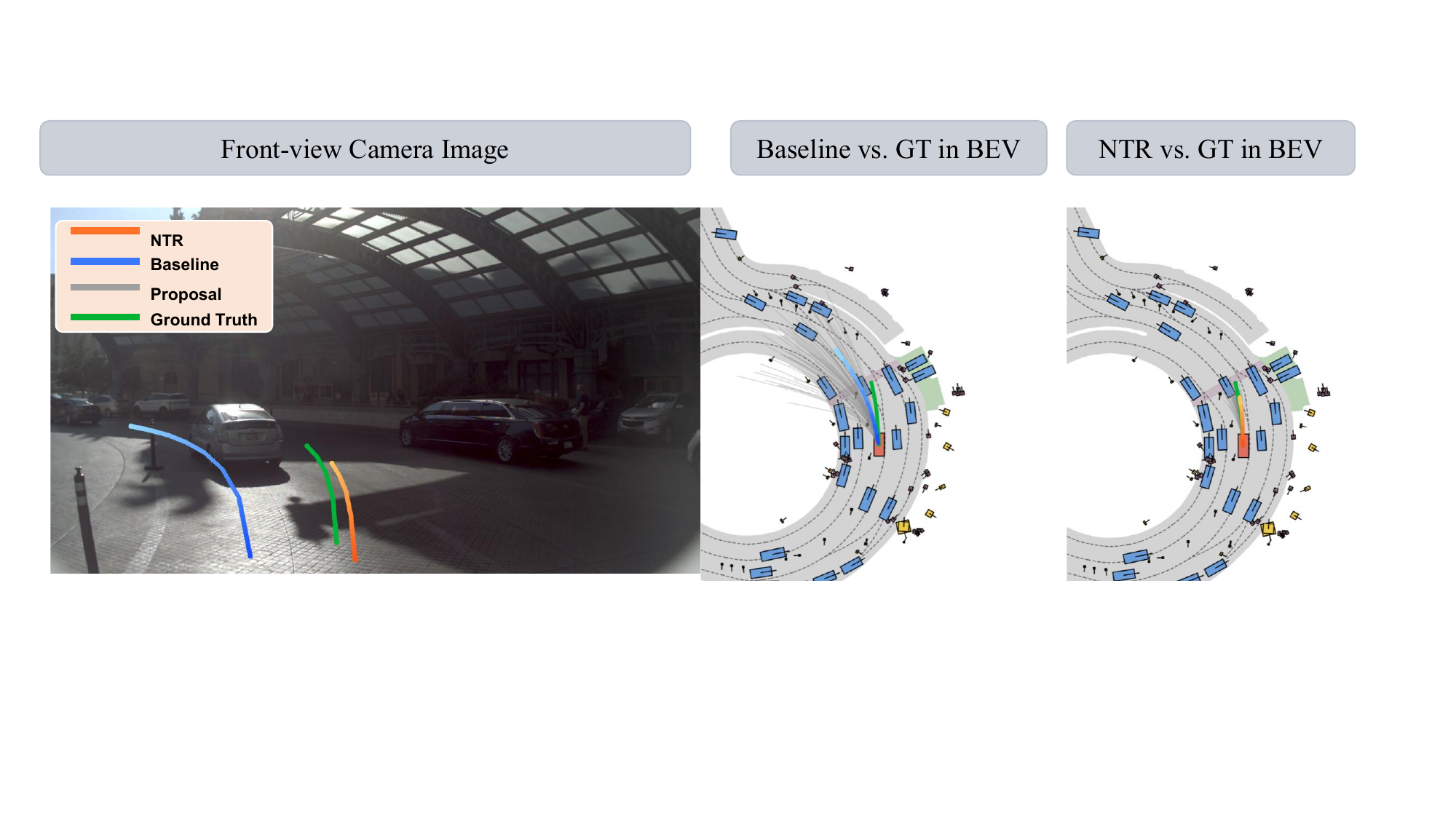}
\caption{
Additional planning trajectory comparison on NavSim. The planning-only baseline is
shown in blue, NTR in orange, and the ground truth in green. 
}
\label{fig:qual_traj}
\end{figure*}

Figure~\ref{fig:qual_traj} shows additional comparisons of the final selected
trajectories for the planning-only baseline and NTR on NavSim. NTR produces
trajectories that are more consistent with the visible road structure and
surrounding context, complementing the quantitative results in the main paper.

\subsection{Failure Cases}
\label{app:qual_fail}

\begin{figure*}[htbp]
\centering
\includegraphics[width=0.8\textwidth,trim={0.8cm 3cm 0.8cm 3cm},clip]{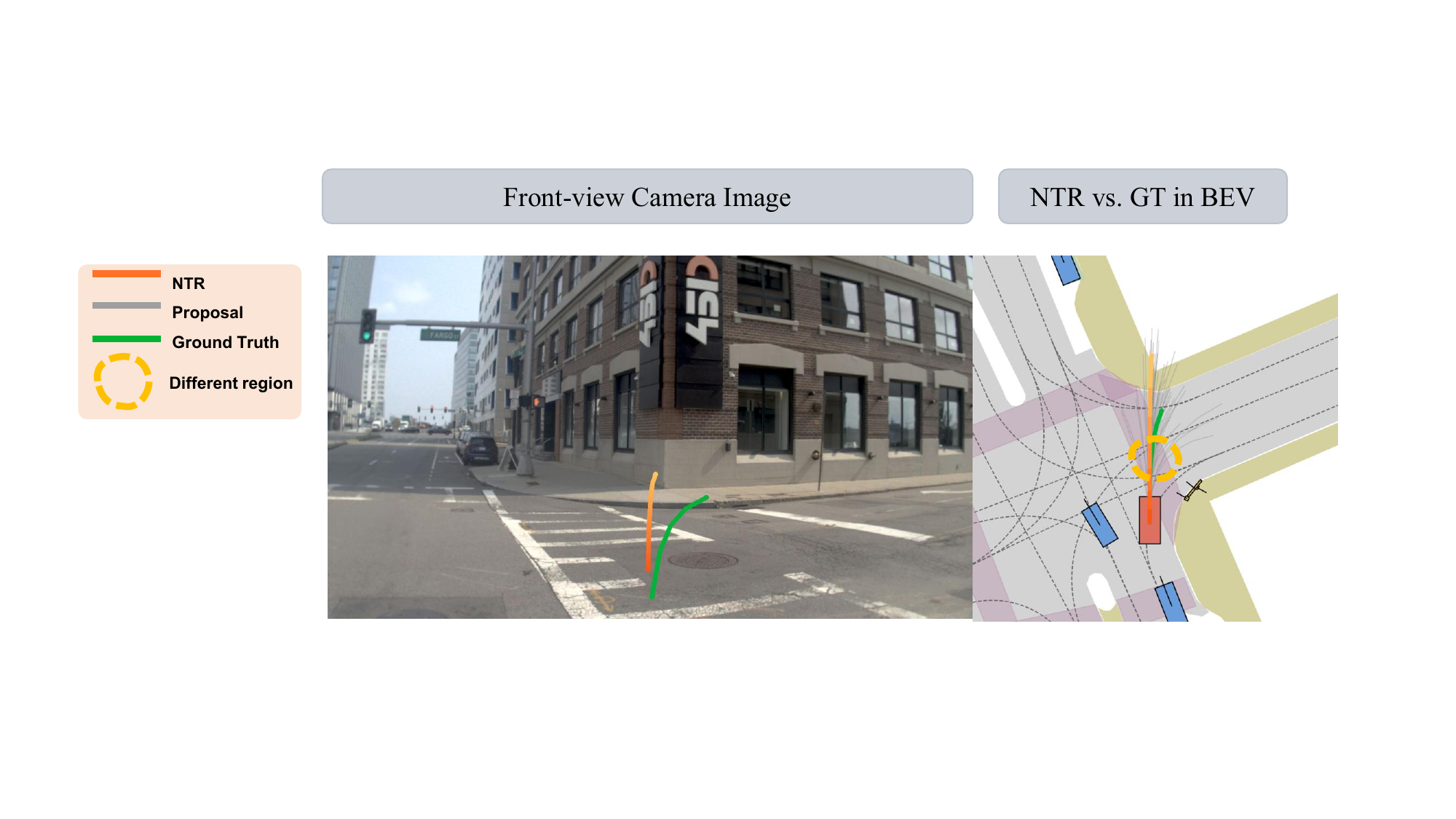}
\vspace{-0.7em}
\includegraphics[width=0.8\textwidth,trim={0.8cm 3cm 0.8cm 3cm},clip]{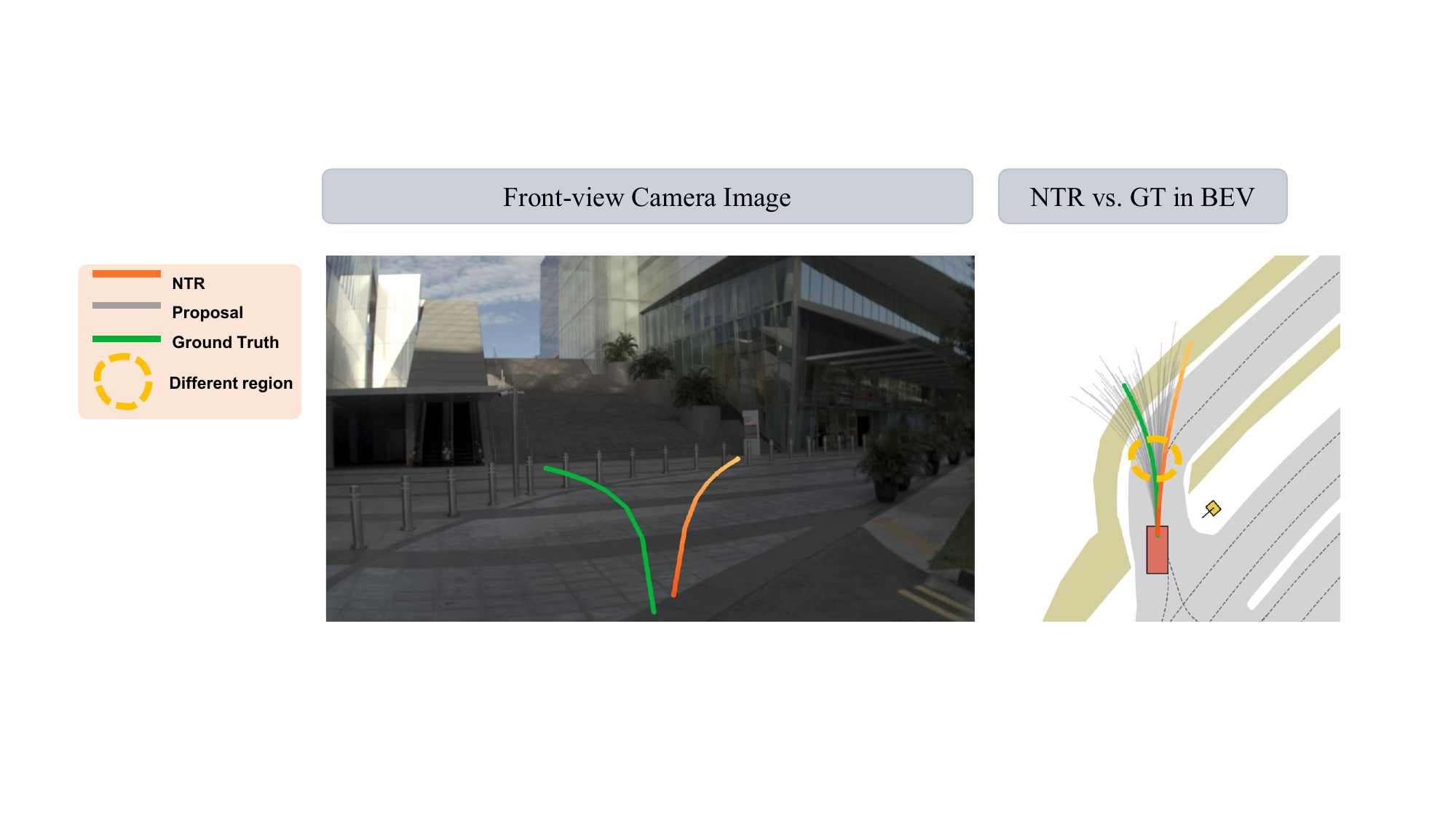}
\vspace{-0.4em}
\caption{
Planning failure examples on NavSim. NTR improves trajectory selection in many
cases but can still deviate from the ground truth under ambiguous intention.
}
\label{fig:planning_failure_navsim}
\end{figure*}

\begin{figure*}[htbp]
\centering
\includegraphics[width=0.8\textwidth,trim={0.8cm 3cm 0.8cm 3cm},clip]{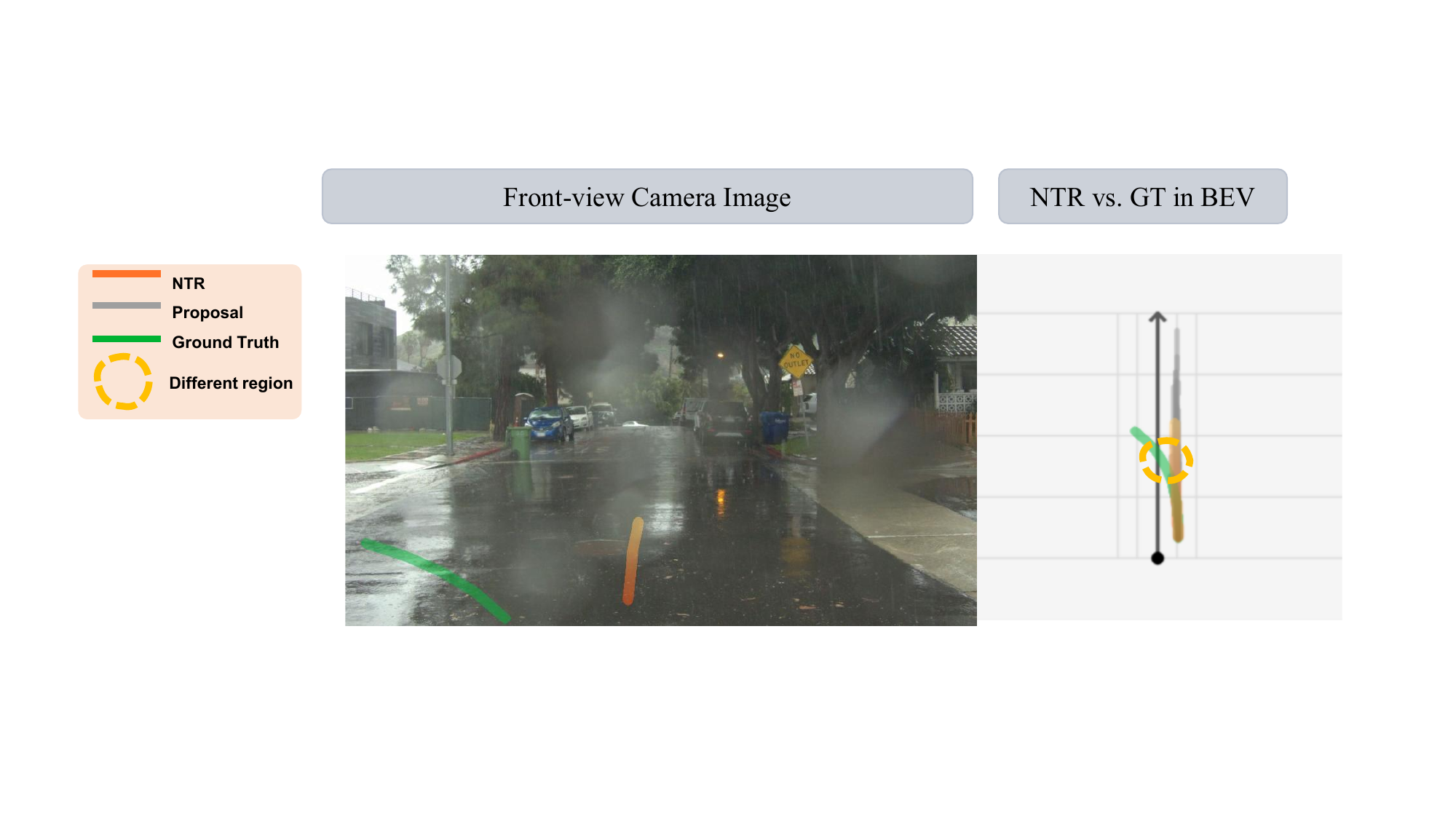}
\vspace{-0.4em}
\caption{
Planning failure examples on Waymo. Even with improved scene-token
representations, the selected trajectory can still deviate from the ground truth
under unclear road geometry and challenging weather conditions.
}
\label{fig:planning_failure_waymo}
\end{figure*}

\paragraph{Planning failure (GT vs.\ NTR).}
Figures~\ref{fig:planning_failure_navsim} and~\ref{fig:planning_failure_waymo}
show cases where NTR's selected trajectory deviates from the ground truth.
Bottleneck supervision improves the scene-token memory but does not resolve every scene ambiguity. The remaining
failures involve ambiguous driving intention, unclear road geometry or challenging weather conditions, and do not show a consistent pattern of missed small actors or long-tail traffic controls.
Moreover, the consistent gains on Waymo E2E, NavSim v1/v2 and our large-scale private dataset suggest that the improvement is not specific to a single benchmark metric.

\begin{figure*}[htbp]
\centering

\includegraphics[width=\textwidth,trim={0.4cm 1.2cm 0.4cm 1.2cm},clip]{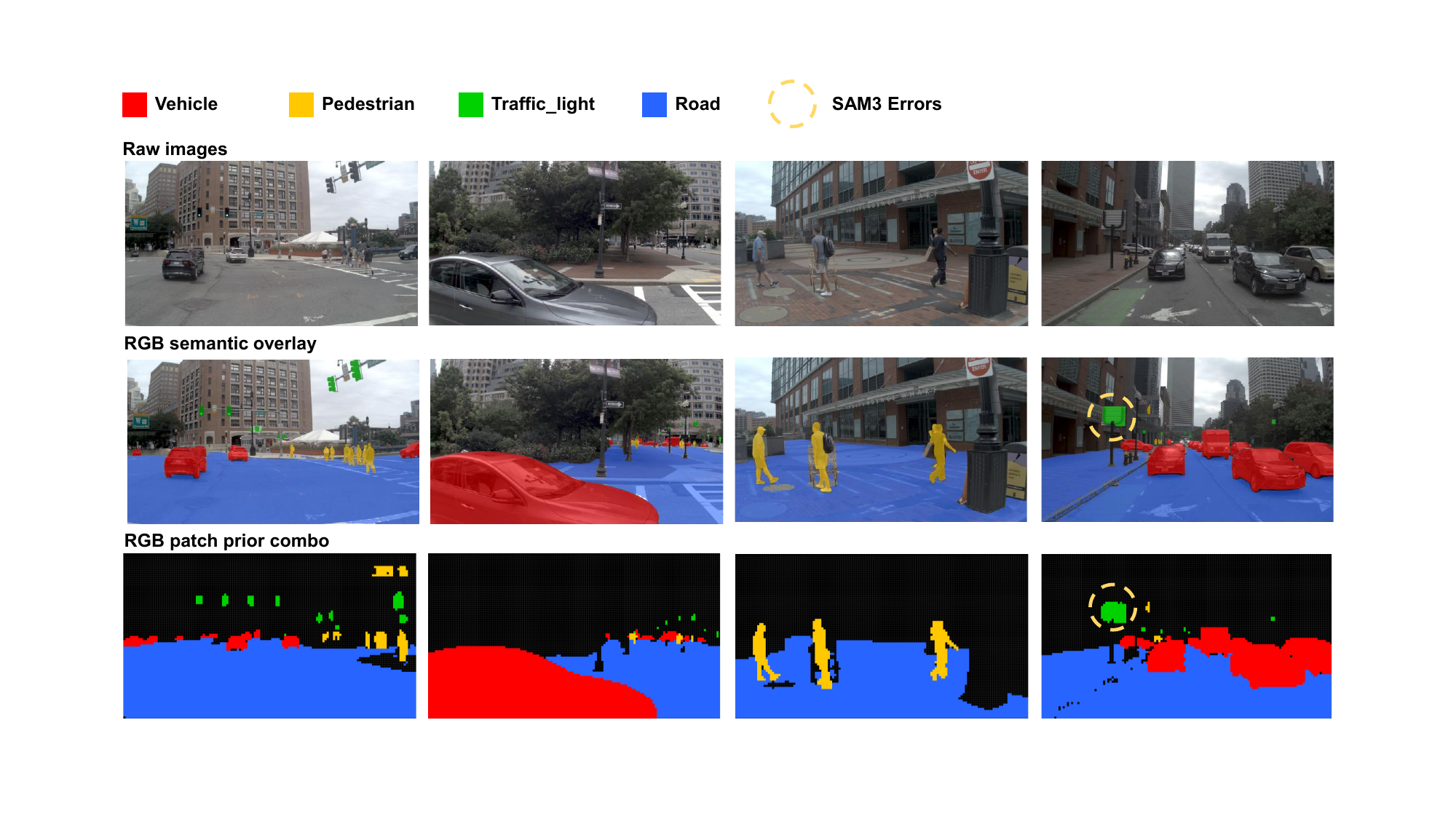}
\par\vspace{-1.2em}

\includegraphics[width=\textwidth,trim={0.4cm 1.2cm 0.4cm 1.2cm},clip]{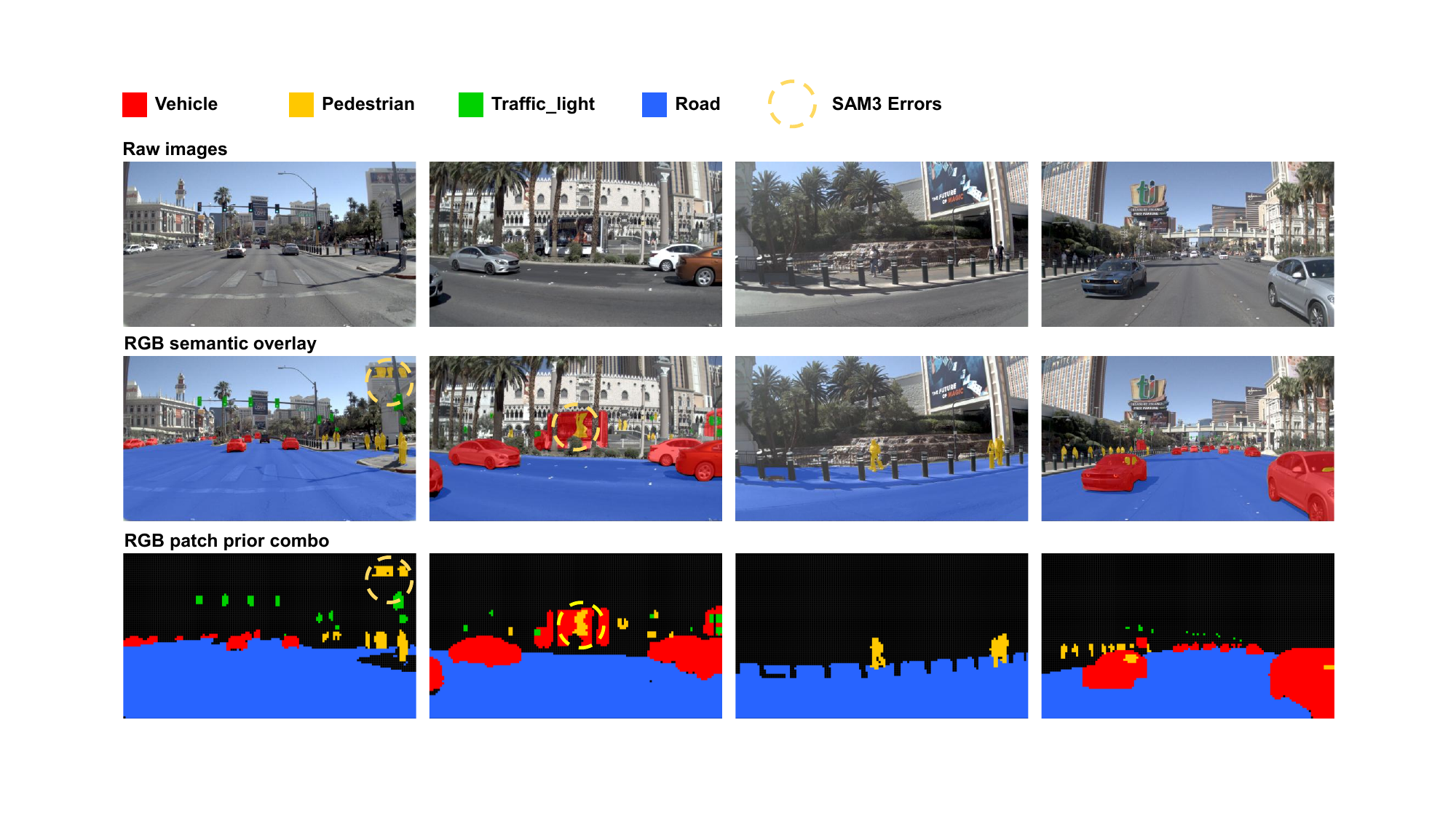}
\par\vspace{-1.2em}

\includegraphics[width=\textwidth,trim={0.4cm 1.2cm 0.4cm 1.2cm},clip]{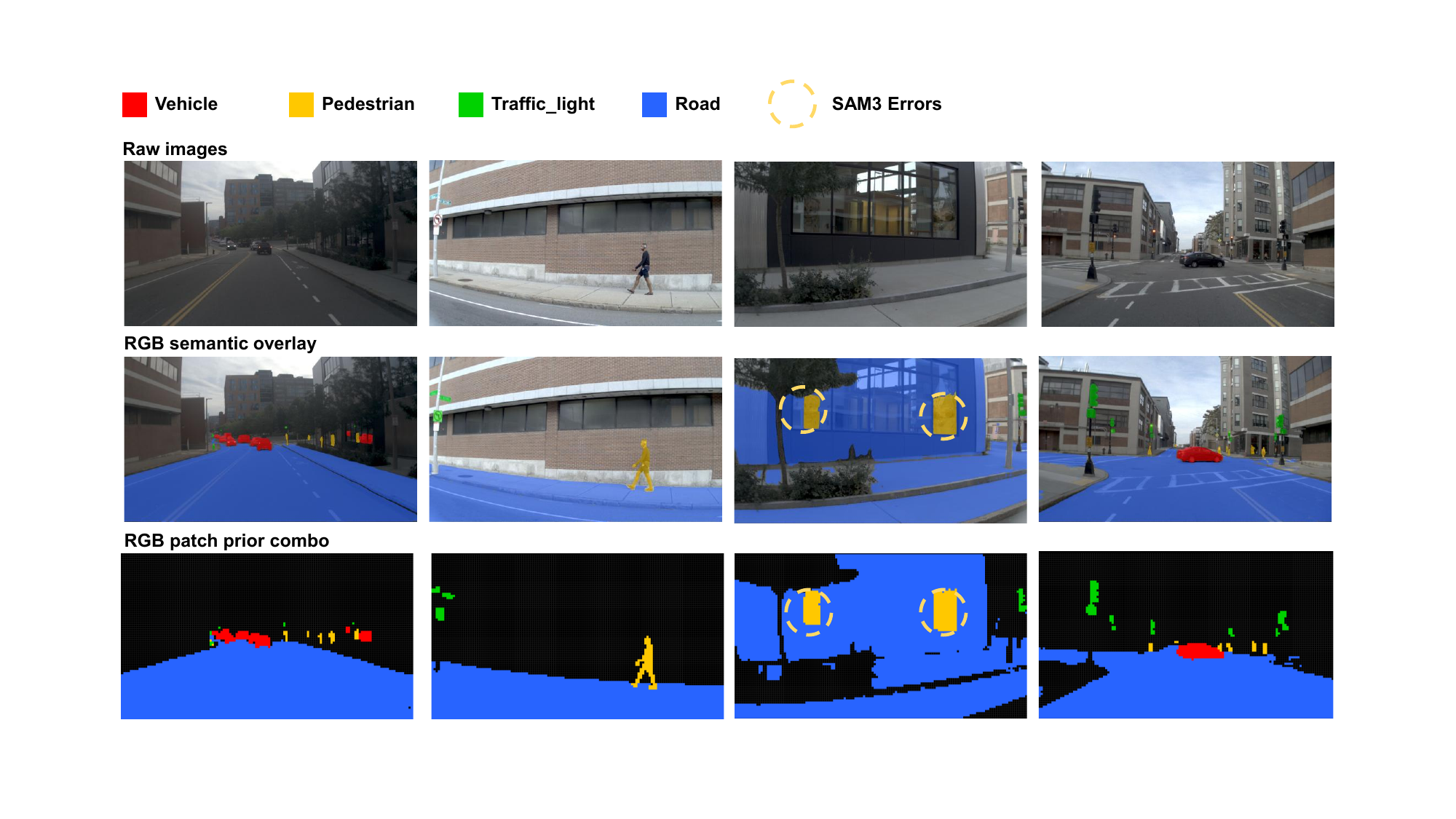}
\par

\caption{
Typical failure cases of SAM3-generated semantic priors.
Top: some road signs or visually salient roadside objects may be confused with traffic-control cues.
Middle and Bottom: human figures appearing in advertisements or posters may be incorrectly recognized as real pedestrians.
Although such noise may affect training-time reconstruction-target allocation, the semantic-prior branch is removed at inference and does not produce perception outputs during deployment.
}
\label{fig:fail_sam3}
\end{figure*}

\paragraph{Annotation failure of semantic priors.}
Figure~\ref{fig:fail_sam3} shows typical failure cases of the SAM3-generated semantic priors.
Although the priors bias reconstruction toward structured driving regions, the generated masks can still be noisy in challenging scenes.
Typical errors include confusing human figures in advertisements with real pedestrians, and mistaking visually salient roadside objects for traffic-control cues.
Since the semantic priors are used only for soft reconstruction-target allocation during training, such noise affects only the quality of the auxiliary supervision and does not introduce perception outputs at inference time.




\end{document}